\documentclass{article}

\usepackage[main, final]{neurips_2026}

\usepackage[utf8]{inputenc}
\usepackage[T1]{fontenc}
\usepackage{hyperref}
\usepackage{url}
\usepackage{booktabs}
\usepackage{amsfonts}
\usepackage{nicefrac}
\usepackage{microtype}

\usepackage[table, svgnames, dvipsnames]{xcolor}
\definecolor{highlight}{HTML}{ffe5ec}
\definecolor{baseline}{HTML}{ececec}
\definecolor{featurecol}{HTML}{dceae5}
\definecolor{downstreamcol}{HTML}{dce4ef}
\definecolor{featurelabel}{HTML}{2a7f7f}
\definecolor{downstreamlabel}{HTML}{2a4d7f}

\usepackage{subcaption}
\usepackage{wrapfig}
\usepackage{graphicx}
\usepackage{tabularx}
\usepackage{multirow}
\usepackage{array, makecell}
\usepackage{rotating}
\usepackage{tabu}

\newcolumntype{P}[1]{>{\centering\arraybackslash}p{#1}}

\newcommand{\rup}[1]{{\tiny\itshape\color{DarkGreen!75}$\uparrow$ #1}}
\newcommand{\rdn}[1]{{\tiny\itshape\color{DarkRed!75}$\downarrow$ #1}}
\newcommand{\rupL}[1]{{\tiny\itshape\color{DarkRed!75}$\uparrow$ #1}}
\newcommand{\rdnL}[1]{{\tiny\itshape\color{DarkGreen!75}$\downarrow$ #1}}

\usepackage{pifont}
\usepackage{bbding}

\usepackage{amsmath}
\usepackage{amssymb}
\usepackage{mathtools}
\usepackage{amsthm}
\usepackage{bbm}
\usepackage{bm}

\title{Breaking Feature Collapse in Self-Supervised Time Series Encoders}

\author{%
  Özgün Turgut\textsuperscript{\ensuremath{1,2}}
  \quad
  Daniel Rueckert\textsuperscript{\ensuremath{2,3,4}}\vspace{.2cm} \\
  \textsuperscript{\ensuremath{1}}{TaktAI}\\
  \textsuperscript{\ensuremath{2}}{Chair for AI in Healthcare and Medicine, Technical University of Munich, Germany}\\
  \textsuperscript{\ensuremath{3}}{Munich Center for Machine Learning, Munich, Germany}\\
  \textsuperscript{\ensuremath{4}}{Department of Computing, Imperial College London, United Kingdom}\\
  \{\texttt{oezguen.turgut, daniel.rueckert}\}\texttt{@tum.de}
}

\begin{document}

\maketitle

\begin{abstract}
General-purpose time series encoders promise a single backbone that generalises across tasks and domains; however, they currently do not scale.
We trace this down to a critical point of failure: under the predominant masked data modelling supervision, the encoder's output tokens \textit{partially collapse}.
This is the natural consequence of supervising encoders only in the input space, which leaves no constraint against redundant token dimensions in the feature space.
We introduce \texttt{OTISv2}, which suppresses the collapse by complementing the reconstruction objective in input space with two objectives in the feature space: a self-distillation loss against an exponential-moving-average teacher, and a KoLeo regulariser.
Under matched training, the resulting encoder reduces the cosine similarity and lifts the effective rank of its output tokens relative to its predecessor, and it scales monotonically from $7.1\,$M to $116\,$M parameters where its predecessor declines. 
It matches or surpasses existing general-purpose time series encoders across $143$ uni- and multi-variate datasets.
To foster further research, we release our code and pre-trained weights\footnote{\url{https://github.com/oetu/otisv2}}.
\end{abstract}

\section{Introduction}

\begin{wrapfigure}[17]{r}{0.35\linewidth}
    \centering
    \vspace{-44pt}
    \includegraphics[width=\linewidth]{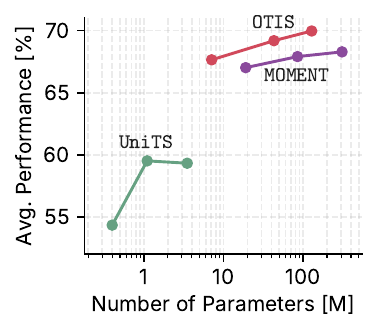}
    \caption{\textbf{General-purpose time series encoders saturate at scale.}
    Average accuracy over $23$ UEA and $120$ UCR datasets.
    Across three general-purpose encoder families, gains through parameter scaling saturate. \texttt{OTIS-7.1M} nearly matches the $43\times$ larger \texttt{MOMENT-308M}.}
    \label{fig:scaling}
\end{wrapfigure}

In recent years, general-purpose backbones have transformed natural language processing~(NLP) and computer vision~(CV).
In CV, for instance, the \texttt{DINO} family \citep{Caron2021, Oquab2024, Simeoni2025} provides dense features that transfer, without retraining, to classification, segmentation, depth estimation, and retrieval across natural, satellite, and medical imaging.
Time series analysis is now undergoing the same transition: \textit{general-purpose encoders} like \texttt{UniTS}~\citep{Gao2024}, \texttt{MOMENT} \citep{Goswami2024}, and \texttt{OTIS} \citep{Turgut2026} have been proposed as backbones that can serve various tasks across domains as diverse as electrocardiography (ECG), electroencephalography (EEG), audio, and weather.
These backbones are pre-trained via \textit{masked data modelling}: patches of their input are hidden, they learn to reconstruct them, and dense features that transfer are assumed to follow.
The most effective family on this frontier is currently \texttt{OTIS}, which matches much larger families using only $7.1\,$M parameters by explicitly tailoring tokenisation, masking, and the reconstruction objective to time series properties.

\textbf{Observation.}\hspace{3mm}
On closer inspection, however, general-purpose time series encoders currently \textit{do not scale}.
At the frontier, \texttt{OTIS-7.1M} comes within $0.6$ percentage points (pp) of \texttt{MOMENT-308M} in downstream performance across $143$ datasets, despite being $43\times$ smaller; a tiny encoder rivalling a much larger one, the opposite of the scaling trend observed in NLP and CV.
Within each family the picture is the same: \texttt{OTIS} gains only $0.8\,$pp from $43\,$M to $128\,$M parameters, \texttt{MOMENT} $0.4\,$pp from $85\,$M to $308\,$M, and \texttt{UniTS} even loses $0.2\,$pp from $1.1\,$M to $3.5\,$M (Fig.~\ref{fig:scaling}).
A universal backbone should improve noticeably with capacity; current general-purpose time series encoders do not.

\begin{figure}[!t]
    \centering
    \includegraphics[width=\linewidth]{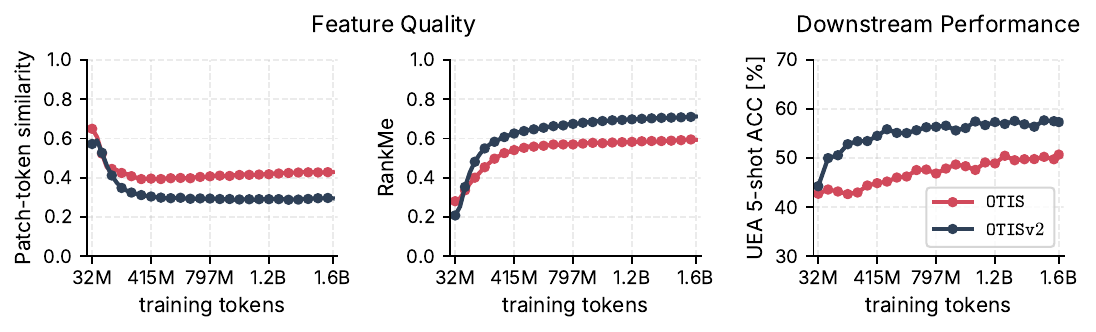}
    \caption{\textbf{Feature quality and its consequences for downstream performance.}
    Pre-training of \texttt{OTIS} and \texttt{OTISv2} with the same data, width, depth, optimiser, and schedule.
    Both use $C\!=\!6ND\!=\!6.82\!\times\!10^{16}$ FLOPs at $N\!=\!7.1\,$M parameters, $D\!=\!1.60\,$B training tokens, and $D_C\!=\!32\,$M unique tokens.
    \textbf{(Left)}~Patch-token cosine similarity~($\downarrow$).
    \textbf{(Centre)}~RankMe~($\uparrow$) normalised by the encoder width $d\!=\!192$.
    \textbf{(Right)}~UEA $5$-shot balanced accuracy on $23$ datasets~($\uparrow$).
    Token similarity drops in both encoders within the first $200\,$M tokens, then stays at $0.46$ in \texttt{OTIS} but declines to $0.32$ in \texttt{OTISv2}, whose $5$-shot lead grows to $5$ to $7\,$pp by $1.60\,$B tokens.}
    \label{fig:teaser}
\end{figure}

\textbf{Analysis.}\hspace{3mm}
We trace this scaling failure down to the feature space: the encoder's \textit{output tokens partially collapse}, being highly similar to one another and spanning only a fraction of the available dimensions (Fig.~\ref{fig:teaser}).
This is a natural consequence of supervising the encoder only in input space: masked data modelling rewards per-patch reconstruction accuracy but places no explicit constraint on the \textit{feature} space, so nothing prevents the encoder from routing information through a small set of token dimensions.
The resulting cap on the effective rank contributes to the scaling plateau, and the few surviving dimensions are easy for specialised encoders to match on narrow benchmarks.

\textbf{Solution.}\hspace{3mm}
To prevent this feature collapse, we add two feature-space objectives that \textit{operate directly on the encoder} and its output tokens, complementing the existing reconstruction objective on the decoder:
(1) a \textit{self-distillation} loss in the spirit of \texttt{DINO} \citep{Caron2021} and \texttt{iBOT} \citep{Zhou2022}, which matches student output tokens to those of an exponential-moving-average teacher, supplying a rich signal that explicitly shapes the feature space;
and (2) a \textit{KoLeo} regulariser \citep{Sablayrolles2019} following \texttt{DINOv2} \citep{Oquab2024}, which spreads the \texttt{[CLS]} features of different samples over the unit hypersphere and thus penalises near-duplicate features across samples.
The result is an objective that now supervises both the input space and feature space of an encoder, which is a prerequisite for positive scaling effects.

\textbf{Outcomes.}\hspace{3mm}
This yields \texttt{OTISv2}, a $7.1\,$M encoder which matches or surpasses general-purpose encoders up to $43\times$ larger across $23$ multi-variate UEA \citep{Bagnall2018} and $120$ uni-variate UCR \citep{Dau2019} datasets (Sec.~\ref{sec:main_comparison});
whose token similarity falls and whose effective rank rises relative to its predecessor under matched training (Sec.~\ref{sec:breaking_feature_collapse});
and which converts parameters ranging from $7.1\,$M to $116\,$M into downstream performance gains, where its predecessor declines (Sec.~\ref{sec:scaling}).

\section{Related Work}
\label{sec:related}

\paragraph{Self-Supervised Learning for Time Series.}
Early time series work relied on \textit{contrastive learning}, aligning augmented views of a sample in the feature space \citep{Kiyasseh2021, Zhang2022, Woo2022, Wang2022, Turgut2025}, but identifying effective augmentations is highly domain-dependent and limits reach across heterogeneous corpora.
\textit{Masked data modelling} has therefore emerged as the dominant paradigm: by reconstructing masked segments, current methods such as \texttt{PatchTST} \citep{Nie2023}, \texttt{UniTS} \citep{Gao2024}, \texttt{MOMENT} \citep{Goswami2024}, and \texttt{OTIS} \citep{Turgut2026} learn features without hand-crafted augmentations.
However, their training signal lives entirely in the \textit{input space}, supervising the encoder only \textit{indirectly} towards semantically consistent features.

\paragraph{Supervision in Feature Space.}
In computer vision, general-purpose encoders are directly supervised in the \textit{feature space}: \texttt{DINO} \citep{Caron2021} supervises a student's \texttt{[CLS]} token features to match the output distribution of an exponential-moving-average teacher, \texttt{iBOT} \citep{Zhou2022} extends this training signal to patch tokens, and \texttt{DINOv2} \citep{Oquab2024} combines the two at scale and adds a KoLeo regulariser \citep{Sablayrolles2019} to spread features over the unit hypersphere.
\texttt{OTISv2} brings these feature-space objectives to time series, motivated by concrete failure modes of existing general-purpose time series encoders.

\paragraph{General-Purpose Backbones for Time Series.}
Recent research aims to unify time series analysis under a single backbone.
\texttt{UniTS} \citep{Gao2024} uses task prompts within a $3.5\,$M backbone, risking learning fixed prompt-to-task mappings rather than generalisable features.
\texttt{MOMENT} \citep{Goswami2024} pairs a $308\,$M encoder with a linear $8\,$k decoder, a capacity mismatch that may weaken the training signal.
\texttt{OTIS} \citep{Turgut2026} reaches competitive performance at only $7.1\,$M parameters, but requires domain labels at tokenisation.

\paragraph{Scaling and Feature Collapse.}
Progress in time series analysis has been driven by the \textit{scaling laws} hypothesis \citep{Kaplan2020, Hoffmann2022}, increasing the parameter counts of backbones to address the heterogeneity of multi-domain time series data.
Feature collapse under self-supervised learning is a known cause of stalled scaling in other modalities, and the remedies that have proven effective at scale act directly on the encoder's output: teacher-based centring \citep{Caron2021, Zhou2022, Oquab2024, Simeoni2025} and entropy-maximising spreads such as KoLeo \citep{Sablayrolles2019, Oquab2024, Simeoni2025}.
We show that the same pathology manifests in self-supervised learning for time series and contributes to the observed scaling saturation in general-purpose time series encoders.

\section{Methodology}
\label{sec:method}
To build \texttt{OTISv2}, we start from the \texttt{OTIS} recipe (Sec.~\ref{sec:arch}) and extend it in two ways.
First, to remove the dependency on domain labels at tokenisation, we replace its domain-specific variate embeddings and absolute position embeddings with register tokens and axial RoPE (Sec.~\ref{sec:tokeniser}).
Second, to counteract the partial collapse of output tokens under masked data modelling, we add a self-distillation loss and a KoLeo regulariser that also supervise the encoder in feature space (Sec.~\ref{sec:mdm}).

\subsection{Architecture}
\label{sec:arch}
\texttt{OTISv2} builds on the Transformer \citep{Vaswani2017} encoder and decoder of \texttt{OTIS} \citep{Turgut2026}.
The encoder $f(\cdot)$ has $7.1\,$M parameters, $12$ blocks of width $d\!=\!192$, three heads, and a $768$-wide MLP.
The decoder~$g(\cdot)$ has $1.7\,$M parameters and is used only during pre-training, with four blocks of width $160$, five heads, and a $640$-wide MLP.
Each variate is split into non-overlapping patches of $p\!=\!24$ time points, so a sample with $V$ variates and $\overline{T}$ time points yields $N_p\!=\!V \cdot T'$ tokens, where $T'\!=\!\overline{T}/p$.
The pre-training context spans $\overline{T}\!=\!1{,}008$ time points, which gives $T'\!=\!42$ patches per variate.

Two changes act inside every block.
A SwiGLU MLP \citep{Shazeer2020} replaces the GELU MLP of \texttt{OTIS}.
The attention and MLP output projections are zero-initialised, so each block starts as the identity \citep{Goyal2017, Oquab2024}.

\subsection{Tokenisation}
\label{sec:tokeniser}
\texttt{OTIS} adds two position embeddings to each patch token.
A fixed sinusoidal embedding encodes time, and a learnable embedding encodes variates \textit{within a domain}.
The learnable variate embedding thus ties the tokeniser to domain labels.
\texttt{OTISv2} removes this dependency, so no domain label is needed at any stage.
To let the encoder store global information outside the patch tokens, we instead add register tokens \citep{Darcet2024} to the input sequence.
The patch tokens can thus retain the local waveform information. 
Axial rotary position embedding (RoPE) \citep{Su2024, Heo2024} replaces both absolute position embeddings, so that position enters attention only through the relative offset between patches.

\paragraph{Register Tokens.}
\label{sec:registers}
Vision Transformers \citep{Dosovitskiy2021} can repurpose patch tokens to store and process global information, at the cost of the local information these tokens carry.
Register tokens \citep{Darcet2024} provide dedicated slots for such global computation.
We include them to separate global from local information.
The registers can hold information about the sample or its domain, while the patch tokens are free to describe the local waveform structure.
This separation matters for \texttt{OTISv2} in particular, since it has no domain-specific variate embeddings to carry such information.
We append $R\!=\!4$ learnable register tokens, denoted \texttt{[reg]}, to the input sequence after patch embedding, alongside the \texttt{[CLS]} token.
Like the \texttt{[CLS]} token, they are initialised from $\mathcal{N}(0, 0.02^2)$ and have no grid position.
At the output, they are discarded, so downstream tasks utilise only \texttt{[CLS]} and patch tokens.

\paragraph{Axial Rotary Position Embedding.}
\label{sec:axial_rope}
Rotary position embedding (RoPE) \citep{Su2024} rotates each query and key vector by angles that depend on its absolute position.
For a query and key $\mathbf{q}_n, \mathbf{k}_{n'} \in \mathbb{R}^{d_h}$ of an attention head at $1$D integer positions $n, n'$, it groups consecutive dimensions into pairs, treats the $j$-th pair as a complex number with frequency $\omega_j$, and maps $q_{n,j} \mapsto q_{n,j}\, e^{\mathrm{i} n\omega_j}$ and $k_{n',j} \mapsto k_{n',j}\, e^{\mathrm{i} n'\omega_j}$.
The attention logit, proportional to $\mathrm{Re}\bigl[\sum_j q_{n,j}\, \overline{k_{n',j}}\, e^{\mathrm{i}(n-n')\omega_j}\bigr]$, depends only on the \textit{relative} offset $n-n'$.
Axial RoPE~\citep{Heo2024} extends this to $2$D inputs by applying RoPE along each axis on a separate half of the head dimension.
It makes the encoder independent of the sequence length and the variate count.
A relative encoding of time lets the encoder process series of any length, without interpolating position embeddings at inference.
Likewise, a relative encoding of the variate index distinguishes the variates of a sample without tying them to a domain, and it extends to any number of variates.
With $d_h\!=\!64$, the first half of the dimensions rotate with the variate index $v \in \{0, \dots, V\!-\!1\}$ of a patch and the last half with its time index $t \in \{0, \dots, T'\!-\!1\}$.
Each half thus has $32$ dimensions, or $16$ pairs, with frequencies $\omega_j\!=\!\Omega^{-j/16}$ for $j\!=\!0, \dots, 15$ and the base $\Omega\!=\!2{\times}10^{5}$.
The \texttt{[CLS]} and \texttt{[reg]} tokens have no grid position and are not rotated.

\subsection{Pre-Training}
\label{sec:mdm}
Following \texttt{OTIS}, we pre-train the encoder via masked data modelling.
Specifically, a subset of patches is masked, the encoder processes only the visible patches, and a decoder reconstructs all patches in input space.
We keep the dual masking strategy \citep{Turgut2026} of \texttt{OTIS}, which applies post-fix masking with probability $\pi\!=\!0.33$ and random masking otherwise.
We use the masking ratio $\rho_s\!=\!0.7$ for random masking and $\rho_s\!=\!0.5$ for post-fix masking, and perturb both with uniform noise from $\mathcal{U}(0, 0.15)$.
Let $\mathbf{x}_i,\widehat{\mathbf{x}}_i \in \mathbb{R}^{1\times p}$ denote the $i$-th input patch and its reconstruction, respectively.
The reconstruction loss optimised is the mean squared error between reconstruction and input,
\begin{equation}
    \mathcal{L}_{\text{MSE}} = \frac{1}{N_p\,p}\sum_{i=1}^{N_p} \| \mathbf{x}_i - \widehat{\mathbf{x}}_i \|_2^2 \; \in \; [0, \infty).
\end{equation}
Additionally, a shape mismatch between reconstruction and input is penalised with a normalised cross-correlation (NCC) regulariser.
For two multi-variate time series $\mathbf{X}, \widehat{\mathbf{X}} \in \mathbb{R}^{V \times \overline{T}}$ it is defined as
\begin{equation}
    \mathcal{L}_{\text{NCC}} = 1 - \mathrm{NCC}(\mathbf{X}, \widehat{\mathbf{X}}) \; \in \; [0, 2],
    \quad
    \mathrm{NCC}(\mathbf{X}, \widehat{\mathbf{X}}) = \frac{1}{V\,\overline{T}}\sum_{v=1}^{V}\sum_{t=1}^{\overline{T}} \frac{x_{v,t} - \mu_v}{\sigma_v}\,\frac{\widehat{x}_{v,t} - \widehat{\mu}_v}{\widehat{\sigma}_v},
\end{equation}
where $\mu_v, \sigma_v$ and $\widehat{\mu}_v, \widehat{\sigma}_v$ denote the mean and standard deviation of variate $v$ in $\mathbf{X}$ and $\widehat{\mathbf{X}}$ over its $\overline{T}$ time points.
Both objectives, $\mathcal{L}_{\text{MSE}}$ and $\mathcal{L}_{\text{NCC}}$, are inherited from \texttt{OTIS} and supervise the encoder in input space. 
The following two objectives add supervision in feature space.

\paragraph{Self-Distillation.}
\label{sec:self_distillation}
Inspired by \texttt{DINO} \citep{Caron2021} and \texttt{iBOT} \citep{Zhou2022}, we add self-distillation to supervise both the \texttt{[CLS]} and patch tokens of the encoder.
A student $f_{\theta_s}$ and a teacher $f_{\theta_t}$ share the encoder architecture.
The teacher receives no gradients and its weights are an exponential moving average (EMA) of the student weights, $\theta_t \leftarrow m\,\theta_t + (1-m)\,\theta_s$, where momentum $m$ follows a cosine schedule from $0.996$ to $1$.
Given a sample, the student processes two independently masked views $u, u'$ with masking ratio $\rho_s$ and sees only the visible patches.
The decoder $g$ reconstructs view $u$ only.
The teacher processes less masked views with masking ratio $\rho_t \leq \rho_s$, whose visible patches are a superset of the student's.
Both $f_{\theta_s}$ and $f_{\theta_t}$ have two projection heads, one each for the \texttt{[CLS]} and patch tokens, each with $K\!=\!8{,}192$ outputs.
We denote them by $h_s$ and $h_t$ and mark the token type of their distributions with the superscript \texttt{[CLS]} or patch.
Like $f_{\theta_t}$, $h_t$ is an EMA of $h_s$ with the same momentum~$m$.
For an output token $\mathbf{z} \in \mathbb{R}^d$ of the respective encoder, the student and teacher distributions are
\begin{equation}
  \mathbf{p}_s = \mathrm{softmax}\bigl(h_s(\mathbf{z}) / \tau_s\bigr) \; \in \; \mathbb{R}^K,
  \qquad
  \mathbf{p}_t = \mathrm{softmax}\bigl((h_t(\mathbf{z}) - \mathbf{c}) / \tau_t\bigr) \; \in \; \mathbb{R}^K,
\end{equation}
where $\mathbf{c}$ is an EMA of the teacher outputs $h_t(\mathbf{z})$ with momentum $0.9$, kept separately for \texttt{[CLS]} and patch tokens.
Centring by $\mathbf{c}$ prevents one dimension from dominating, and a teacher temperature $\tau_t < \tau_s\!=\!0.1$ prevents collapse to the uniform distribution.
Sample-level and patch-level losses are
\begin{equation}
    \mathcal{L}_{\text{CLS}} = -\tfrac{1}{2}\bigl[\mathbf{p}_t^{\texttt{[CLS]}}(u')^{\top} \log \mathbf{p}_s^{\texttt{[CLS]}}(u) + \mathbf{p}_t^{\texttt{[CLS]}}(u)^{\top} \log \mathbf{p}_s^{\texttt{[CLS]}}(u')\bigr] \; \in \; [0, \infty),
\end{equation}
\begin{equation}
    \mathcal{L}_{\text{patch}} = -\tfrac{1}{2}\sum_{w \in \{u, u'\}} \frac{1}{|\mathcal{V}_w|}\sum_{i \in \mathcal{V}_w} \mathbf{p}_{t,i}^{\text{patch}}(w)^{\top} \log \mathbf{p}_{s,i}^{\text{patch}}(w) \; \in \; [0, \infty),
\end{equation}
where $\mathcal{V}_w$ indexes the visible patches of view $w$, and $\mathbf{p}_t(w)$ is computed from the less masked teacher view of $w$.
The $\mathcal{L}_{\text{CLS}}$ term pulls the sample-level representations of different views together.
The $\mathcal{L}_{\text{patch}}$ term instead pulls each visible patch token of the student towards the teacher token at the same position, which the teacher computes from the same or more context.
This gives dense supervision in feature space that masked data modelling in input space cannot provide.

\paragraph{KoLeo Regularisation.}
\label{sec:koleo}
Self-distillation reduces \textit{intra-sample} redundancies.
On its own, it does not prevent \textit{inter-sample} redundancies, since the student can still map different inputs to a narrow cone in feature space.
We therefore add a KoLeo regulariser \citep{Sablayrolles2019}, which spreads features uniformly over the unit hypersphere.
It builds on the Kozachenko--Leonenko estimator of differential entropy.
Given $\ell_2$-normalised student \texttt{[CLS]} tokens $\mathbf{z}_1, \dots, \mathbf{z}_{n_g} \in \mathbb{R}^d$ of a group of $n_g\!=\!32$ masked samples,
\begin{equation}
    \mathcal{L}_{\text{KoLeo}} = -\frac{1}{n_g}\sum_{i=1}^{n_g} \log \bigl(r_{i}\bigr) \; \in \; [-\log 2, \infty),
    \qquad
    r_{i} = \min_{j \in \{1, \dots, n_g\} \setminus \{i\}} \|\mathbf{z}_i - \mathbf{z}_j\|_2,
\end{equation}
rewards a larger distance $r_i$ from each sample $i$ to its nearest neighbour $j$ and thus directly penalises near-duplicate features.
We average this loss over all groups of a batch.

\paragraph{Training Objective.}
\label{sec:total_loss}
The full pre-training objective of \texttt{OTISv2} combines the input-space objectives of \texttt{OTIS} with the feature-space objectives introduced above,
\begin{equation}
    \mathcal{L} \;=\; \mathcal{L}_{\text{MSE}}
    \;+\; \lambda_{\text{NCC}}\,\mathcal{L}_{\text{NCC}}
    \;+\; \lambda_{\text{SD}}\,\bigl(\mathcal{L}_{\text{CLS}} + \mathcal{L}_{\text{patch}}\bigr)
    \;+\; \lambda_{\text{KoLeo}}\,\mathcal{L}_{\text{KoLeo}}.
    \label{eq:total_loss}
\end{equation}
We empirically set $\lambda_{\text{NCC}}\!=\!0.1$, $\lambda_{\text{SD}}\!=\!0.1$, and $\lambda_{\text{KoLeo}}\!=\!0.025$.
All losses are averaged over the batch.

\subsection{Data}
\label{sec:data}
We curate a diverse corpus of $756\,$k samples and $45\,$B time points (App.~\ref{sec:app_ts_corpus}) covering $11$ domains across healthcare (ECG, EEG), audio, engineering (electromechanics), natural sciences (temperature, weather), infrastructure (cloud, energy, traffic), and finance (banking, economics).
This breadth exposes \texttt{OTISv2} to heterogeneous variate correlations and temporal dynamics.
Variate counts range from $1$ to $19$, and sequence lengths from $720$ to $480\,$k time points.
To align with the growing interest in deploying general-purpose encoders for healthcare applications \citep{Saab2024, Zhang2024}, our pre-training corpus includes the most widely used physiological time series.
In particular, it contains $400\,$k $12$-lead ECGs ($24\,$B time points) and $3.4\,$k $10$--$20$ system EEG recordings ($3.5\,$B time points).

\section{Experiments}
\label{sec:experiments}
We pre-train \texttt{OTISv2} for $50$ epochs on two NVIDIA A$100$ GPUs with AdamW \citep{Loshchilov2017}, effective batch size $384$, peak learning rate $1.2{\times}10^{-3}$ with a warmup-stable-decay schedule \citep{Hu2024}, weight decay $0.025$, layer-wise learning rate decay $0.975$, drop path $0.1$ \citep{Huang2016}, and gradient clipping at $3$.
Compute costs and hyperparameters are detailed in Appendix~\ref{sec:app_comp_cost}.

We evaluate \texttt{OTISv2} along five axes.
First, we compare its downstream performance on $143$ datasets from the UEA \citep{Bagnall2018} and UCR~\citep{Dau2019} archives (App.~\ref{sec:app_benchmark_details}) against nine general-purpose encoders (App.~\ref{sec:app_baseline_details}, Sec.~\ref{sec:main_comparison}).
Second, we track the feature quality of \texttt{OTIS} and \texttt{OTISv2} during matched pre-training to test whether feature-space supervision mitigates feature collapse (Sec.~\ref{sec:breaking_feature_collapse}).
Third, we vary feature quality under a fixed objective to test whether it determines downstream performance (Sec.~\ref{sec:dose_response}).
Fourth, we conduct a cumulative ablation from \texttt{OTIS} to \texttt{OTISv2} to attribute the changes in feature quality and downstream performance to the architecture, the training stabilisers, and the objective (Sec.~\ref{sec:ablation}).
Fifth, we scale the encoder from $7.1\,$M to $116\,$M parameters under data constraints to test whether mitigating feature collapse eventually unlocks positive scaling effects (Sec.~\ref{sec:scaling}).

\paragraph{Metrics.}
\label{sec:metrics}
We pair downstream performance with two feature quality measures (App.~\ref{sec:app_exp_details}).
\textit{Patch-token cosine similarity} (or \textit{token similarity} for short, $\downarrow$) is the average pairwise cosine similarity between the $\ell_2$-normalised output patch tokens of a single time series, quantifying token redundancy within a sample.
\textit{RankMe} \citep{Garrido2023} ($\uparrow$) measures the effective rank of the mean-pooled patch tokens, reflecting the dimensionality spanned across samples.
Bounded by the encoder width $d$, RankMe is compared only across encoders of the same width.
Both are measured on frozen checkpoints, except in Sections~\ref{sec:breaking_feature_collapse} and~\ref{sec:scaling}, which record them during training and compare them only within the same figure (App.~\ref{sec:app_exp_details}).
Downstream performance is measured via \textit{balanced accuracy}~($\uparrow$) under three evaluation protocols: $5$-shot prototypical classification, $k$-NN classification, and linear probing.
\textit{Avg} is the balanced accuracy averaged over all datasets and evaluation protocols, such that each dataset carries equal weight.

\subsection{Downstream Performance of Frozen General-Purpose Encoders}
\label{sec:main_comparison}
\paragraph{Setup.}
We evaluate how well time series encoders generalise to unseen datasets without fine-tuning.
Table~\ref{tab:main} compares \texttt{OTISv2-7.1M} with nine public encoders of the \texttt{UniTS}, \texttt{MOMENT}, and \texttt{OTIS} families.
As these encoders were trained under different conditions, including training data, token budgets, and compute, this comparison serves solely as a benchmark. 
Complementarily, Sections~\ref{sec:breaking_feature_collapse} to~\ref{sec:scaling} test how our proposed architectural changes and objectives affect pre-training under matched conditions.

\begin{table}[!t]
  \caption{\textbf{Feature quality and downstream performance of frozen general-purpose time series encoders.}
  Balanced accuracy on $23$ UEA and $120$ UCR datasets, with \textbf{best} and \underline{second-best} scores highlighted.
  $D$ counts patch tokens seen during pre-training, including repetition.
  \texttt{OTISv2-7.1M} matches or surpasses encoders up to $43\times$ larger on average.}
  \label{tab:main}
  \centering
  \footnotesize
  \setlength{\tabcolsep}{0.45em}
  \begin{tabular}{lrllllllllll l}
      \toprule
       & & & & \multicolumn{2}{>{\columncolor{featurecol}}c}{\textit{\color{featurelabel}Feature Quality}} & \multicolumn{7}{>{\columncolor{downstreamcol}}c}{\textit{\color{downstreamlabel}Downstream Performance}} \\[2pt]
       Encoder
       & \rotatebox{90}{\shortstack[l]{Encoder\\width $d$}}
       & \rotatebox{90}{\shortstack[l]{Compute\\$C\!=\!6ND$\\{\tiny[FLOPs]}$^{\ddagger}$}}
       & \rotatebox{90}{\shortstack[l]{Tokens/param\\$\mathrm{TPP}\!=\!D/N$}}
       & \cellcolor{featurecol}\rotatebox{90}{\shortstack[l]{Patch-token\\cosine sim {\tiny($\downarrow$)}}}
       & \cellcolor{featurecol}\rotatebox{90}{RankMe {\tiny($\uparrow$)}}
       & \cellcolor{downstreamcol}\rotatebox{90}{\shortstack[l]{UEA\\{\scriptsize\textit{5-shot}} {\tiny($\uparrow$)}}}
       & \cellcolor{downstreamcol}\rotatebox{90}{\shortstack[l]{UEA\\{\scriptsize\textit{k-NN}} {\tiny($\uparrow$)}}}
       & \cellcolor{downstreamcol}\rotatebox{90}{\shortstack[l]{UEA\\{\scriptsize\textit{linear}} {\tiny($\uparrow$)}}}
       & \cellcolor{downstreamcol}\rotatebox{90}{\shortstack[l]{UCR\\{\scriptsize\textit{5-shot}} {\tiny($\uparrow$)}}}
       & \cellcolor{downstreamcol}\rotatebox{90}{\shortstack[l]{UCR\\{\scriptsize\textit{k-NN}} {\tiny($\uparrow$)}}}
       & \cellcolor{downstreamcol}\rotatebox{90}{\shortstack[l]{UCR\\{\scriptsize\textit{linear}} {\tiny($\uparrow$)}}}
       & \cellcolor{downstreamcol}\rotatebox{90}{\shortstack[l]{Avg\\{\scriptsize\textit{weighted}} {\tiny($\uparrow$)}}} \\
      \midrule
      \texttt{UniTS-0.4M}                                              & 32          & $4.3{\times}10^{16}$$^{\dagger}$ & $45{,}000$$^{\dagger}$ & 0.23             & 16                     & 38.48             & 46.02              & 50.92              & 49.71              & 57.36              & 61.29              & 54.35              \\
      \texttt{UniTS-1.1M}$^{\S}$                                       & 64          & $6.0{\times}10^{16}$$^{\dagger}$ & $8{,}200$$^{\dagger}$ & 0.09             & 32                     & 46.46             & 52.78              & 56.45              & 53.97              & 62.93              & 66.03              & 59.52              \\
      \texttt{UniTS-3.5M}                                              & 128         & $3.8{\times}10^{17}$$^{\dagger}$ & $5{,}200$$^{\dagger}$ & 0.27             & 46                     & 44.82             & 51.77              & 58.17              & 52.92              & 61.43              & 68.10              & 59.33              \\
      \midrule
      \texttt{MOMENT-19M}                                              & 512         & $3.5{\times}10^{16}$$^{\dagger}$ & $16$$^{\dagger}$ & 0.84               & 25                     & 48.13             & 54.77              & 63.60              & 60.87              & 69.41              & 77.41              & 67.02              \\
      \texttt{MOMENT-85M}                                              & 768         & $1.6{\times}10^{17}$$^{\dagger}$ & $3.6$$^{\dagger}$ & 0.83               & 49                     & 49.46             & 56.43              & 63.63              & 62.28              & 70.24              & \underline{77.75}  & 67.91              \\
      \texttt{MOMENT-308M}                                             & 1{,}024     & $5.7{\times}10^{17}$$^{\dagger}$ & $1.0$$^{\dagger}$ & 0.56               & 181                    & 52.84             & 58.37              & 64.38              & \underline{62.80}  & 70.39              & 77.28              & 68.29              \\
      \midrule
      \texttt{OTIS-7.1M}                                               & 192         & $1.8{\times}10^{18}$ & $5{,}700$ & 0.55               & 68                     & 55.53             & 64.42              & 67.73              & 59.36              & 69.88              & 76.63              & 67.65              \\
      \texttt{OTIS-43M}                                                & 384         & $5.1{\times}10^{18}$ & $470$ & 0.67               & 99                     & 57.53              & 64.80              & \underline{69.28}  & 61.45              & 71.52              & 77.66              & 69.19              \\
      \texttt{OTIS-128M}                                               & 576         & $1.6{\times}10^{19}$ & $160$ & 0.51               & 162                    & \underline{58.70}  & \underline{65.78}  & 69.19              & 62.37              & \underline{72.46}  & \textbf{78.18}     & \underline{69.97}  \\
      \arrayrulecolor{lightgray}\midrule\arrayrulecolor{black}
      \rowcolor{highlight}
      \texttt{\textbf{OTISv2-7.1M}}                                    & 192         & $4.6{\times}10^{17}$ & $1{,}500$ & 0.60      & 62            & \textbf{60.00}    & \textbf{66.22}     & \textbf{69.84}     & \textbf{63.20}     & \textbf{72.70}     & 77.18              & \textbf{70.12}     \\
      \bottomrule
      \multicolumn{13}{l}{\footnotesize $^{\dagger}$Not reported by the original works; reconstructed from their released pre-training recipes and configurations (App.~\ref{sec:app_baseline_details}).}\\
      \multicolumn{13}{l}{\footnotesize $^{\ddagger}$$C\!=\!6ND$ \citep{Kaplan2020} is reported for cross-family comparability only (App.~\ref{sec:app_comp_cost}).}\\
      \multicolumn{13}{l}{\footnotesize $^{\S}$Fully supervised, not pre-trained by masked data modelling (App.~\ref{sec:app_baseline_details}).}
  \end{tabular}
\end{table}

\paragraph{Results.}
\texttt{OTISv2-7.1M} matches or surpasses general-purpose encoders up to $43\times$ larger.
With $70.12\,\%$, it yields the best average performance.
Only \texttt{OTIS-128M} comes within $0.2\,$pp ($69.97\,\%$), with $18\times$ more parameters and $1.9\times$ more training tokens.
The largest encoder, \texttt{MOMENT-308M}, stays $1.83\,$pp below despite $43\times$ more parameters and $1.2\times$ more compute.
Over \texttt{OTIS-7.1M} of the same size, \texttt{OTISv2} gains $2.47\,$pp on average and $0.6$ to $4.5\,$pp on every archive and protocol, with $3.8\times$ fewer training tokens.
\texttt{OTIS} gains about as much only by growing from $7.1\,$M to $128\,$M parameters ($2.32\,$pp), so at a fixed size the recipe of \texttt{OTISv2} is worth an $18$-fold increase in parameters.

\texttt{OTISv2} has the highest accuracy under all three protocols on UEA and two of the three on UCR, and its lead is largest under the $5$-shot evaluation protocol ($1.30\,$pp on UEA and $0.40\,$pp on UCR over the next-best encoder).
Its frozen features thus separate classes from few labels, on multi-variate and uni-variate data alike.
Only on the UCR linear probe do five encoders score $0.1$ to $1.0\,$pp higher.
These are exactly the five wider encoders ($d\!=\!384$ to $1{,}024$, against $192$), so here the trained linear head appears to profit from encoder width.

The margin of \texttt{OTISv2} over \texttt{MOMENT} and \texttt{UniTS} grows with the number of variates $V$.
On all eight datasets with $V\!\geq\!8$, \texttt{OTISv2} outperforms \texttt{MOMENT-308M} and \texttt{UniTS-3.5M} and matches \texttt{OTIS-128M}, scoring higher on four.
Every encoder scores lower on these datasets than on uni-variate ones, but the \texttt{OTIS} family loses the least accuracy: \texttt{OTIS-128M} $4.2\,$pp, \texttt{OTISv2} $5.1\,$pp, \texttt{UniTS-3.5M} $9.9\,$pp, and \texttt{MOMENT-308M} $12.7\,$pp.
Four of the eight datasets have $24$ to $144$ variates, more than the $19$ seen in pre-training.
Against \texttt{MOMENT-308M}, the margin also grows with high-frequency content (Spearman $\rho\!=\!0.28$), whereas neither dataset difficulty nor similarity to our corpus predicts it.
The advantage of \texttt{OTISv2} thus holds on data unlike its pre-training data, in both variate count and signal properties.

Under matched training, \texttt{OTISv2} has lower token similarity and higher RankMe than \texttt{OTIS} ($0.50$ against $0.56$ and $75$ against $52$, Sec.~\ref{sec:ablation}).
The released encoders do not show this order ($0.60$ against $0.55$ and $62$ against $68$), as they also differ in training data and training length (App.~\ref{sec:app_baseline_details}).
Token similarity and RankMe thus compare recipes only between runs trained alike.

\subsection{Feature Collapse With and Without Feature-Space Supervision}
\label{sec:breaking_feature_collapse}
\paragraph{Setup.}
\texttt{OTIS} is supervised only in input space, whereas \texttt{OTISv2} also supervises its feature space through self-distillation and KoLeo regularisation.
If this supervision mitigates feature collapse, \texttt{OTISv2} should have lower token similarity and higher RankMe than \texttt{OTIS}.
To test this, we pre-train both encoders from scratch under matched conditions: the same data, token budget, width, depth, optimiser, and schedule.
The training curves are depicted in Figure~\ref{fig:teaser}.
The two encoders still differ in architecture and training stabilisers, which Section~\ref{sec:ablation} separates from the feature-space objectives.

\paragraph{Results.}
\texttt{OTISv2} keeps lowering token similarity after \texttt{OTIS} stalls, and its downstream lead grows with training.
In both encoders, token similarity drops within the first $200\,$M training tokens.
\texttt{OTIS} then stays at $0.46$, whereas \texttt{OTISv2} declines further to $0.32$.
The RankMe of \texttt{OTISv2} also rises faster and levels off higher.
In \texttt{OTIS}, the collapse thus sets in early, and further training does not undo it.
The UEA $5$-shot accuracy of the two encoders is at parity after $32\,$M tokens, and \texttt{OTISv2} leads by $5$ to $7\,$pp after $1.60\,$B tokens.
More training thus widens the lead of \texttt{OTISv2} instead of closing it.

The difference is not limited to the output tokens.
After training, \texttt{OTISv2} has lower token similarity and higher RankMe at every encoder block, and its accuracy lead builds up over the first four blocks (App.~\ref{sec:app_layer_kshot}).
Neither encoder develops the high-norm artefact tokens of large vision transformers \citep{Darcet2024} (App.~\ref{sec:app_token_norms}).
Single ECG samples of the released encoders illustrate the difference: in \texttt{OTISv2}, different patches attend to distinct sets of patches, whereas in \texttt{OTIS} their attention maps look alike (App.~\ref{sec:app_ecg_inspection}).

\subsection{Linking Feature Quality to Downstream Performance}
\label{sec:dose_response}
\paragraph{Setup.}
Whether mitigating feature collapse causes the downstream lead of \texttt{OTISv2} is still open, as the objective and the degree of collapse change together.
We therefore keep the objective fixed and vary the degree of collapse by changing only the optimisation.
If downstream performance then follows, there is a causal link between collapse and performance.
We pre-train \texttt{OTISv2-7.1M} at the ten combinations of effective batch size $B_\text{eff}$ and peak learning rate $\eta$ in Table~\ref{tab:dose_response}, keeping the objective, architecture, data, training token budget $D\!=\!221\,$M (as in Sec.~\ref{sec:scaling}), and compute fixed.

\begin{table}[!t]
  \caption{\textbf{Feature quality and downstream performance under a fixed objective.}
  \texttt{OTISv2-7.1M} pre-trained at ten combinations of effective batch size $B_\text{eff}$ and peak learning rate $\eta$, ordered by RankMe.
  RankMe, patch-token cosine similarity (Sim), and UEA $5$-shot accuracy ($5$-shot) are means over three seeds.
  RankMe correlates with $5$-shot ($r\!=\!+0.83$), but Sim does not ($r\!=\!-0.48$).}
  \label{tab:dose_response}
  \centering
  \footnotesize
  \setlength{\tabcolsep}{0.7em}
  \begin{tabular}{lccc@{\hspace{2em}}lccc}
      \toprule
      $B_\text{eff}$ / $\eta$ & Sim ($\downarrow$) & RankMe ($\uparrow$) & $5$-shot ($\uparrow$) & $B_\text{eff}$ / $\eta$ & Sim ($\downarrow$) & RankMe ($\uparrow$) & $5$-shot ($\uparrow$) \\
      \midrule
      $384$ / $3{\times}10^{-4}$   & $0.528$ & $51.2$ & $52.18$ & $384$ / $1.2{\times}10^{-3}$ & $0.519$ & $62.1$ & $53.22$ \\
      $192$ / $3{\times}10^{-4}$   & $0.487$ & $58.8$ & $53.26$ & $96$ / $3{\times}10^{-4}$    & $\mathbf{0.481}$ & $62.6$ & $54.29$ \\
      $384$ / $2.4{\times}10^{-3}$ & $0.556$ & $59.3$ & $52.99$ & $192$ / $1.2{\times}10^{-3}$ & $0.516$ & $62.9$ & $53.69$ \\
      $384$ / $6{\times}10^{-4}$   & $0.505$ & $59.8$ & $53.00$ & $192$ / $6{\times}10^{-4}$   & $0.489$ & $64.3$ & $54.27$ \\
      $96$ / $1.2{\times}10^{-3}$  & $0.538$ & $59.8$ & $54.09$ & $96$ / $6{\times}10^{-4}$    & $0.494$ & $\mathbf{64.8}$ & $\mathbf{55.01}$ \\
      \bottomrule
  \end{tabular}
\end{table}

\paragraph{Results.}
Downstream accuracy follows the effective rank even under a fixed objective.
Across the ten combinations, RankMe spans $51$ to $65$, and UEA $5$-shot accuracy follows it with a Pearson correlation of $r\!=\!+0.83$.
When the $5$-shot protocol uses the mean-pooled patch tokens instead of the \texttt{[CLS]} token, the correlation is $r\!=\!+0.78$, so the link does not depend on which embedding is classified.
As the objective is fixed, the loss of rank itself limits downstream performance.

Only RankMe tracks accuracy on both data archives.
Token similarity spans only $0.48$ to $0.56$ and does not track UEA $5$-shot accuracy ($r\!=\!-0.48$).
On the uni-variate UCR archive, both measures track $5$-shot accuracy ($r\!=\!+0.66$ with RankMe and $r\!=\!-0.85$ with token similarity), although accuracy varies by only $1.8\,$pp across the combinations.
The run with the highest RankMe, $B_\text{eff}\!=\!96$ at $\eta\!=\!6{\times}10^{-4}$, also has the highest UEA $5$-shot accuracy ($55.01\,\%$).
It takes the most optimisation steps at a moderate learning rate, whereas at every batch size, raising the learning rate beyond the best value lowers both RankMe and accuracy.
At $B_\text{eff}\!=\!384$, which the scaling study of Section~\ref{sec:scaling} holds fixed, the peak learning rate of the released \texttt{OTISv2}, $1.2{\times}10^{-3}$, is the best setting.
As RankMe needs no labels, it can thus guide the choice of optimisation settings before any downstream evaluation.

\subsection{Cumulative Ablation of Architecture, Objective, and Stabilisation}
\label{sec:ablation}
\paragraph{Setup.}
\texttt{OTISv2} changes the architecture, objective, and training stabilisers of \texttt{OTIS}.
To isolate the contribution of each component, we start from the \texttt{OTIS} recipe and add one component per row of Table~\ref{tab:cumulative}, eventually yielding \texttt{OTISv2}.
The first three rows make the architecture independent of domain labels, sequence length, and variate count: shared registers in place of domain-specific variate embeddings, and axial RoPE in place of absolute position embeddings.
The next four rows add training stabilisers: zero-initialised residual projections, SwiGLU MLPs, drop path, and layer-wise learning rate decay.
The last three rows add self-distillation, gradient clipping, and KoLeo regularisation.
If supervising the feature space mitigates feature collapse, only these last three rows should lower token similarity below that of \texttt{OTIS}.

\begin{table}[!t]
  \caption{\textbf{Cumulative ablation from \texttt{OTIS} to \texttt{OTISv2}.}
  One component added per row, starting from the \texttt{OTIS} pre-training recipe towards the \texttt{OTISv2} recipe.
  All rows use $C\!=\!6ND\!=\!6.82\!\times\!10^{16}$ FLOPs at $N\!=\!7.1\,$M parameters, $D\!=\!1.60\,$B training tokens, and $D_C\!=\!32\,$M unique tokens.
  Mean scores over three pre-training seeds are reported.
  Arrows give the change from the previous row.}
  \label{tab:cumulative}
  \centering
  \scriptsize
  \setlength{\tabcolsep}{0.225em}
  \begin{tabular}{llllllllll}
      \toprule
       & \multicolumn{2}{>{\columncolor{featurecol}}c}{\textit{\color{featurelabel}Feature Quality}} & \multicolumn{7}{>{\columncolor{downstreamcol}}c}{\textit{\color{downstreamlabel}Downstream Performance}} \\[2pt]
       Encoder
       & \cellcolor{featurecol}\rotatebox{90}{\shortstack[l]{Patch-token\\cosine sim {\tiny($\downarrow$)}}}
       & \cellcolor{featurecol}\rotatebox{90}{RankMe {\tiny($\uparrow$)}}
       & \cellcolor{downstreamcol}\rotatebox{90}{\shortstack[l]{UEA\\{\scriptsize\textit{5-shot}} {\tiny($\uparrow$)}}}
       & \cellcolor{downstreamcol}\rotatebox{90}{\shortstack[l]{UEA\\{\scriptsize\textit{k-NN}} {\tiny($\uparrow$)}}}
       & \cellcolor{downstreamcol}\rotatebox{90}{\shortstack[l]{UEA\\{\scriptsize\textit{linear}} {\tiny($\uparrow$)}}}
       & \cellcolor{downstreamcol}\rotatebox{90}{\shortstack[l]{UCR\\{\scriptsize\textit{5-shot}} {\tiny($\uparrow$)}}}
       & \cellcolor{downstreamcol}\rotatebox{90}{\shortstack[l]{UCR\\{\scriptsize\textit{k-NN}} {\tiny($\uparrow$)}}}
       & \cellcolor{downstreamcol}\rotatebox{90}{\shortstack[l]{UCR\\{\scriptsize\textit{linear}} {\tiny($\uparrow$)}}}
       & \cellcolor{downstreamcol}\rotatebox{90}{\shortstack[l]{Avg\\{\scriptsize\textit{weighted}} {\tiny($\uparrow$)}}} \\
      \midrule
      \texttt{OTIS-7.1M}                                               & 0.56     & 52             & 52.48             & 56.56              & 64.38              & 58.62              & 61.75              & 74.11              & 63.69              \\
      $+$domain token                                                  & 0.59 \rupL{0.03}    & 51 \rdn{1}              & 51.08 \rdn{1.40}   & 56.78 \rup{0.22}   & 64.31 \rdn{0.07}   & 57.01 \rdn{1.61}   & 60.53 \rdn{1.22}   & 72.56 \rdn{1.55}   & 62.40 \rdn{1.29}   \\
      $+$register tokens                                               & 0.59 \rdnL{0.00}    & 50 \rdn{1}              & 52.78 \rup{1.70}   & 57.60 \rup{0.82}   & 65.51 \rup{1.20}   & 58.37 \rup{1.36}   & 62.04 \rup{1.51}   & 73.58 \rup{1.02}   & 63.69 \rup{1.29}   \\
      $+$Axial RoPE                                                    & 0.66 \rupL{0.07}    & 56 \rup{6}              & 53.12 \rup{0.34}   & 57.97 \rup{0.37}   & 65.98 \rup{0.47}   & 58.75 \rup{0.38}   & 62.26 \rup{0.22}   & 74.83 \rup{1.25}   & 64.27 \rup{0.58}   \\
      $+$zero-init residuals                                           & 0.64 \rdnL{0.02}    & 54 \rdn{2}              & 53.44 \rup{0.32}   & 57.58 \rdn{0.39}   & 66.10 \rup{0.12}   & 56.47 \rdn{2.28}   & 58.77 \rdn{3.49}   & 73.49 \rdn{1.34}   & 62.29 \rdn{1.98}   \\
      $+$SwiGLU                                                        & 0.62 \rdnL{0.02}    & 57 \rup{3}              & 50.41 \rdn{3.03}   & 55.72 \rdn{1.86}   & 64.65 \rdn{1.45}   & 56.50 \rup{0.03}   & 60.60 \rup{1.83}   & 73.28 \rdn{0.21}   & 62.41 \rup{0.12}   \\
      $+$drop path                                                     & 0.64 \rupL{0.02}    & 58 \rup{1}              & 52.26 \rup{1.85}   & 57.20 \rup{1.48}   & 65.58 \rup{0.93}   & 57.99 \rup{1.49}   & 60.62 \rup{0.02}   & 73.75 \rup{0.47}   & 63.19 \rup{0.78}   \\
      $+$layer-wise LR decay                                                & 0.62 \rdnL{0.02}    & 60 \rup{2}              & 52.46 \rup{0.20}   & 56.64 \rdn{0.56}   & 65.35 \rdn{0.23}   & 57.54 \rdn{0.45}   & 61.10 \rup{0.48}   & 73.66 \rdn{0.09}   & 63.14 \rdn{0.05}   \\
      $+$self-distillation                                             & 0.53 \rdnL{0.09}    & 68 \rup{8}              & 55.52 \rup{3.06}   & 59.79 \rup{3.15}   & 67.39 \rup{2.04}   & 59.61 \rup{2.07}   & 63.91 \rup{2.81}   & 75.71 \rup{2.05}   & 65.52 \rup{2.38}   \\
      $+$gradient clipping                                                 & 0.55 \rupL{0.02}    & 67 \rdn{1}              & 55.46 \rdn{0.06}   & 59.55 \rdn{0.24}   & 66.93 \rdn{0.46}   & 59.72 \rup{0.11}   & 63.96 \rup{0.05}   & 75.98 \rup{0.27}   & 65.60 \rup{0.08}   \\
      $+$KoLeo $=$ \texttt{OTISv2-7.1M}                                & 0.50 \rdnL{0.05}    & 75 \rup{8}              & 57.52 \rup{2.06}   & 63.10 \rup{3.55}   & 67.44 \rup{0.51}   & 61.32 \rup{1.60}   & 68.23 \rup{4.27}   & 76.15 \rup{0.17}   & 67.62 \rup{2.02}   \\
      \bottomrule
  \end{tabular}
\end{table}

\paragraph{Results.}
The feature-space objectives supply the entire gain from \texttt{OTIS} to \texttt{OTISv2}.
Together with gradient clipping, they lower token similarity from $0.62$ to $0.50$, raise RankMe from $60$ to $75$, and lift the average accuracy over all datasets and protocols by $4.48\,$pp, more than the total gain of $3.93\,$pp.
The other seven rows together lose $0.55\,$pp: the architectural rows add $0.58\,$pp, and the training stabilisers cost $1.13\,$pp.
Self-distillation lowers token similarity by $0.09$, raises RankMe by $8$, and lifts UEA $5$-shot and $k$-NN accuracy by $3.06$ and $3.15\,$pp.
KoLeo lowers token similarity by another $0.05$, raises RankMe by another $8$, and lifts both accuracies by $2.06$ and $3.55\,$pp.
Although KoLeo acts only on the \texttt{[CLS]} token, it lowers patch-token cosine similarity, as \texttt{[CLS]} and patch tokens share the encoder weights.
Gradient clipping leaves the average accuracy unchanged ($+0.08\,$pp).
As only the last three rows take token similarity below the $0.56$ of \texttt{OTIS}, the lower token similarity of \texttt{OTISv2} in Section~\ref{sec:breaking_feature_collapse} and its downstream gain both stem from supervising the feature space.

The architectural rows make the encoder independent of domain labels, sequence length, and variate count without lowering accuracy.
Shared variate embeddings with a domain token lower the average accuracy by $1.29\,$pp, and registers in place of this token recover the loss ($+1.29\,$pp) at unchanged token similarity.
With registers, the encoder thus applies to any domain and still reaches the same average accuracy as \texttt{OTIS} ($63.69\,\%$).
Axial RoPE raises token similarity more than any other row ($+0.07$), but, among the architectural rows, it alone raises RankMe ($+6$) and yields the largest gain on the UCR linear probe ($+1.25\,$pp).
A rise in token similarity thus need not mean a loss of rank.

\begin{figure}[!t]
    \centering
    \includegraphics[width=\linewidth]{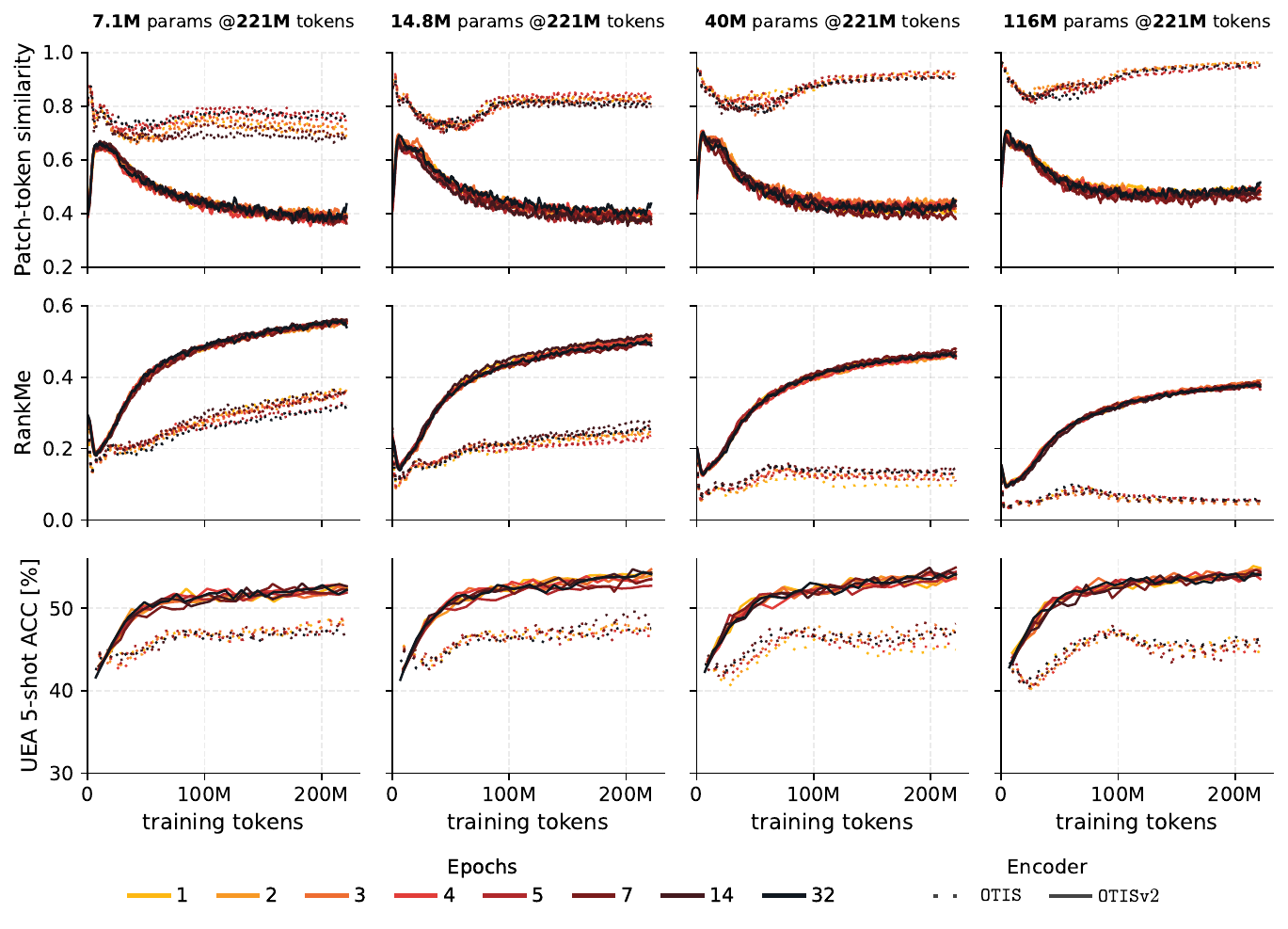}\\[0.4em]
    \small
    \begin{tabular*}{\linewidth}{@{\extracolsep{\fill}}cccc@{}}
        \toprule
        Compute ($C\!=\!6ND$) & Parameters ($N$) & Training tokens ($D$) & Data budget ($D_C$) \\
        \midrule
        $9.4\times 10^{15}$ & $7.1\,\mathrm{M}$  & $221\,\mathrm{M}$ & $\{221,\,111,\,74,\,55,\,44,\,32,\,16,\,7\}\,\mathrm{M}$ \\
        $2.0\times 10^{16}$ & $14.8\,\mathrm{M}$ & $221\,\mathrm{M}$ & $\{221,\,111,\,74,\,55,\,44,\,32,\,16,\,7\}\,\mathrm{M}$ \\
        $5.3\times 10^{16}$ & $40\,\mathrm{M}$   & $221\,\mathrm{M}$ & $\{221,\,111,\,74,\,55,\,44,\,32,\,16,\,7\}\,\mathrm{M}$ \\
        $1.5\times 10^{17}$ & $116\,\mathrm{M}$  & $221\,\mathrm{M}$ & $\{221,\,111,\,74,\,55,\,44,\,32,\,16,\,7\}\,\mathrm{M}$ \\
        \bottomrule
    \end{tabular*}
    \caption{\textbf{Scaling encoder size under data constraints.}
    \textbf{Columns.} The four IsoFLOP $(N,D)$ groups listed in the table.
    \textbf{Rows.} \textbf{(Top)} Patch-token cosine similarity ($\downarrow$). 
    \textbf{(Centre)} RankMe ($\uparrow$) normalised by the encoder width $d$.
    \textbf{(Bottom)} UEA $5$-shot balanced accuracy on $23$ datasets, using mean-pooled patch tokens~($\uparrow$).
    Within each column, the eight data budgets $D_C$ correspond to $\{1,2,3,4,5,7,14,32\}$ epochs of repetition (colour). 
    \texttt{OTIS} is dotted, \texttt{OTISv2} is solid. 
    At every size and data budget, \texttt{OTISv2} has lower token similarity, higher RankMe, and higher accuracy than \texttt{OTIS}.
    }
    \label{fig:efficiency}
\end{figure}
\subsection{Scaling Encoder Size Under Data Constraints}
\label{sec:scaling}
\paragraph{Setup.}
Larger \texttt{OTIS} encoders improve little over smaller ones (Fig.~\ref{fig:scaling}).
If feature collapse causes this plateau, mitigating it should let larger encoders convert their parameters into downstream performance.
As time series corpora are small, Figure~\ref{fig:efficiency} compares both encoders across sizes under data constraints.
We train both encoders at four sizes $N$ (App.~\ref{sec:app_encoder_variants}) on the same training token budget $D\!=\!221\,$M, so that compute is fixed within each size.
At each size, we vary the unique data budget $D_C$ in eight steps from $221\,$M to $7\,$M tokens, i.e.\ from $1$ to $32$ epochs of repetition.
As $D$ is the same for every size, tokens per parameter fall from $31$ at $7.1\,$M to $1.9$ at $116\,$M.
Both encoders use the same settings at every size: effective batch size $384$, peak learning rate $1.2{\times}10^{-3}$, and a warmup-stable-decay schedule.
We approximate compute as $C\!=\!6ND$ \citep{Kaplan2020, Muennighoff2023}, which understates the cost of \texttt{OTISv2} by $1.05$ to $2.1\times$ over our size range (App.~\ref{sec:app_comp_cost}).
Each run uses a single seed.
From four runs at $7.1\,$M repeated with three seeds, the UEA $5$-shot difference between two runs varies by about $0.6\,$pp across seeds, and we read only effects beyond this seed band.

\paragraph{Results.}
\texttt{OTISv2} gains with size where \texttt{OTIS} declines.
At all $32$ combinations of size and data budget, \texttt{OTISv2} has lower token similarity, higher RankMe, and higher UEA $5$-shot accuracy than \texttt{OTIS}, with a margin of $3.6$ to $10.3\,$pp.
Averaged over the eight data budgets, the UEA $5$-shot accuracy of \texttt{OTISv2} rises at every step in size, from $52.3\,\%$ at $7.1\,$M to $54.3\,\%$ at $116\,$M, whereas that of \texttt{OTIS} falls at every step, from $47.9\,\%$ to $45.5\,\%$.
The gap between them 
thus doubles, from $4.4$ to $8.8\,$pp.

\texttt{OTIS} collapses as it grows: its token similarity rises from $0.72$ to $0.95$, and its RankMe, normalised by the encoder width $d$, falls from $0.35$ to $0.05$.
\texttt{OTISv2} loses some rank ($0.55$ to $0.38$), but its similarity stays between $0.39$ and $0.49$.
Both encoders see nearly the same tokens per parameter at every size, so undertraining alone does not explain the collapse of \texttt{OTIS}.
As the loss of rank limits downstream performance (Sec.~\ref{sec:dose_response}), this collapse is the likely cause of the decline of \texttt{OTIS} with size.

Most of the accuracy gain of \texttt{OTISv2} comes with the first doubling of parameters.
From $7.1\,$M to $14.8\,$M, \texttt{OTISv2} improves at all eight budgets beyond the seed band ($+0.8$ to $+2.5\,$pp), whereas \texttt{OTIS} improves at only two.
Beyond $14.8\,$M, the gains of \texttt{OTISv2} are small and change sign between data budgets, but both larger encoders stay above the $7.1\,$M encoder at all eight budgets.
At $40\,$M, \texttt{OTIS} instead scores below its own $14.8\,$M encoder at seven of the eight budgets, and at $116\,$M below its own $7.1\,$M encoder at all eight.
The released \texttt{OTIS} encoders gain $3.2\,$pp in UEA $5$-shot accuracy from $7.1\,$M to $128\,$M parameters (Tab.~\ref{tab:main}), but with $91\times$ to $185\times$ our $221\,$M training tokens.
Larger \texttt{OTISv2} encoders thus pay off on far fewer tokens than larger \texttt{OTIS} encoders.
As the gains beyond $14.8\,$M stay small at this token budget, the plateau sets in at a larger size rather than disappearing.

Repeating data up to $32$ times does not lower the UEA $5$-shot accuracy of \texttt{OTISv2}.
At each size, the mean accuracy of the \texttt{OTISv2} runs with $14$ and $32$ epochs differs from that of runs with $1$ and $2$ epochs by only $-0.5$ to $+0.1\,$pp, inside the seed band, against $-1.4$ to $+1.0\,$pp for \texttt{OTIS}.
As $D$ and the number of optimisation steps are the same at each size, the comparison isolates repetition itself.
For language models, in contrast, the value of repeated tokens decays beyond about four epochs \citep{Muennighoff2023}.
At this scale, $7\,$M unique tokens repeated $32$ times thus serve about as well as $221\,$M unique tokens.

\section{Conclusion}
\label{sec:conclusion}
In this work, we identify feature collapse as a proximate cause of the scaling plateau in general-purpose time series encoders.
We complement the input-space objectives of masked data modelling with two feature-space objectives: a self-distillation loss and a KoLeo regulariser.
This mitigates the collapse, and a single frozen \texttt{OTISv2-7.1M} matches or surpasses general-purpose encoders up to $43\times$ larger in average accuracy over $143$ UEA and UCR datasets.
Our analysis yields three insights.
First, the loss of effective rank limits downstream performance, as accuracy follows the effective rank even under a fixed objective.
Second, under matched training, the feature-space objectives supply the entire accuracy gain over \texttt{OTIS}, whereas the architectural changes mainly make the encoder independent of domain labels, sequence length, and variate count.
Third, once the collapse is mitigated, larger encoders gain accuracy, whereas larger \texttt{OTIS} encoders lose it.
At our token budget, however, the plateau sets in at a larger size rather than disappearing.

\paragraph{Limitations.}
Our evaluation focuses on discriminative tasks, and beyond zero-shot forecasting (App.~\ref{sec:app_forecasting}), the effect of \texttt{OTISv2} on generative tasks remains to be studied.
Furthermore, the pre-training corpus, although diverse, over-represents physiological signals, so generalisation to unseen time series domains needs further evaluation.
The absolute scale we explore, at most $116\,$M parameters and $221\,$M training tokens in the scaling study, is small compared to computer vision and natural language processing.
Whether the recipe extends to billion-parameter time series encoders, and where the plateau sets in at larger token budgets, remain open.
The patch width $p\!=\!24$ limits performance on very short uni-variate series of $96$ time points or fewer, where the encoder sees at most four tokens.
\texttt{OTISv2} is also more sensitive than \texttt{OTIS} to additive noise and, less so, to spikes, missing segments, and variate dropout, in part because its \texttt{[CLS]} embedding, which carries most of the downstream advantage, is more sensitive than its mean-pooled patch tokens (App.~\ref{sec:app_corruption}).
Finally, self-distillation with the exponential-moving-average teacher raises the memory needed for training by about $2.4\times$, which limits batch and encoder size on a given GPU, but it does not affect inference cost.

\begin{ack}
We thank Laura Kraus for encouraging the submission of this work and for believing in its potential. 
\end{ack}

\bibliography{neurips_2026.bib}
\bibliographystyle{abbrv}


\clearpage


\appendix

\section{Pre-Training Corpus Details}
\label{sec:app_ts_corpus}
Table~\ref{tab:app_ts_corpus} summarises the pre-training corpus of \texttt{OTISv2}.
The corpus holds $756{,}267$ samples with $45.2$ billion time points from $11$ domains.
More than half of the samples ($53\,\%$) are physiological signals, namely $400\,$k $12$-lead ECGs and $3.4\,$k EEGs in the $10$--$20$ system, to support general-purpose encoders for healthcare applications \citep{Saab2024, Zhang2024}.
In terms of time points, ECG ($24.0\,$B) and audio ($15.6\,$B) account for $88\,\%$ of the corpus.
Samples have $1$ to $19$ variates and $720$ to $480\,$k time points, covering a wide range of correlations between variates and of temporal dynamics.
The nine sources other than GIFT-Eval \citep{Aksu2024} are shared with \texttt{OTIS} and pre-processed in the same way \citep{Turgut2026}.
The four GIFT-Eval domains are new to \texttt{OTISv2} and are drawn from the GIFT-Eval pre-training collection.
For these domains, we discard series shorter than $1{,}024$ time points, split longer ones into chunks of at most $100\,$k time points ($50\,$k for energy and weather), set missing values to zero, normalise each variate to zero mean and unit variance, and clip the result to $[-10, 10]$.

\begin{table}[!ht]
\caption{\textbf{Pre-training corpus.} Eleven domains from healthcare, audio, engineering, natural sciences, infrastructure, and finance, with $756{,}267$ samples and $45.2\,$B time points in total.
$V\cdot T$ counts the time points of all variates, summed over samples, and Frequency denotes the sampling rate.}
\label{tab:app_ts_corpus}
\centering
\small
\setlength{\tabcolsep}{0.3em}
\begin{tabular}{llrrrrr}
    \toprule
    \textbf{Domain} & \textbf{Source} & \textbf{Samples} & \textbf{Variates} $\bm{V}$ & \textbf{Time points} $\bm{T}$ & $\bm{V \cdot T}$ & \textbf{Frequency} \\
    \midrule
    ECG               & MIMIC-IV-ECG \citep{Gow2023}        & $400{,}000$ & $12$     & $5{,}000$                & $24.0\,$B  & $500\,$Hz \\
    \multirow{2}{*}{EEG}               & TDBRAIN \citep{VanDijk2022}         & $2{,}692$   & $19$     & $60{,}000$               & $3.07\,$B  & $500\,$Hz \\
    & SEED \citep{Zheng2015}              & $675$       & $19$     & $37{,}000$               & $475\,$M   & $200\,$Hz \\
    Audio             & AudioSet-$20$K \citep{Gemmeke2017}  & $19{,}614$  & $1$--$2$ & $24{,}576$--$480{,}000$  & $15.6\,$B  & $44.1\,$kHz \\
    Electromechanics  & FD-A \citep{Lessmeier2016}          & $8{,}184$   & $1$      & $5{,}120$                & $41.9\,$M  & $64\,$kHz \\
    Temperature       & DWD \citep{DWD2024}                 & $202{,}712$ & $1$      & $720$                    & $146\,$M   & $278\,\mu$Hz \\
    Cloud             & GIFT-Eval \citep{Aksu2024}          & $29{,}273$  & $1$--$2$ & $1{,}025$--$8{,}064$      & $167\,$M   & $0.28$--$3.33\,$mHz \\
    Energy            & GIFT-Eval \citep{Aksu2024}          & $21{,}854$  & $1$--$3$ & $1{,}036$--$50{,}000$     & $1.24\,$B  & $0.28$--$250\,$mHz \\
    Traffic           & GIFT-Eval \citep{Aksu2024}          & $67{,}984$  & $1$--$3$ & $1{,}037$--$96{,}400$     & $378\,$M   & $0.28$--$3.33\,$mHz \\
    Weather           & GIFT-Eval \citep{Aksu2024}          & $3{,}053$   & $1$--$11$ & $1{,}302$--$50{,}000$    & $87.8\,$M  & $12$--$278\,\mu$Hz \\
    Banking           & NN5 \citep{Taieb2012}               & $111$       & $1$      & $791$                    & $87.8\,$k  & $12\,\mu$Hz \\
    \multirow{2}{*}{Economics}         & FRED-MD \citep{McCracken2016}       & $107$       & $1$      & $728$                    & $77.9\,$k  & $386\,$nHz \\
    & Exchange \citep{Lai2018}            & $8$         & $1$      & $7{,}588$                & $60.7\,$k  & $12\,\mu$Hz \\
    \midrule
    \textbf{Total}        &                                                    & $756{,}267$ &  &  & $45.2\,$B &  \\
    \bottomrule
\end{tabular}
\end{table}

\section{Computational Details}
\label{sec:app_comp_cost}
Table~\ref{tab:app_pt_cost} reports the compute used for pre-training, and Table~\ref{tab:app_pt_hparams} lists the pre-training hyperparameters of \texttt{OTISv2-7.1M}.
\texttt{OTISv2-7.1M} is pre-trained on two NVIDIA A$100$-$80$GB GPUs.
The $14.8\,$M, $40\,$M, and $116\,$M variants are trained only in the scaling study of Section~\ref{sec:scaling}, each on a single GPU.
Their entries in Table~\ref{tab:app_pt_cost} thus report the total cost of the eight runs per size, one per data budget.
Throughout the paper, we approximate compute as $C\!=\!6ND$.
Counting the matrix multiplications and convolutions of a full training step with the FLOP counter of PyTorch, for a sample of average size ($288$ patch tokens) and at the mean masking ratios of the released recipe, the actual cost of \texttt{OTISv2} per training token is $2.09$, $1.52$, $1.19$, and $1.05$ times $6N$ for the $7.1\,$M, $14.8\,$M, $40\,$M, and $116\,$M encoders.
The decoder, the projection heads, and the teacher add compute that $6ND$ ignores, whereas the encoder processes only the visible patches, so the gap shrinks as the encoder grows.
Each downstream evaluation runs on a single GPU with at least $46\,$GB of memory.

\begin{table}[!ht]
\caption{\textbf{Computational resources used for pre-training \texttt{OTISv2}.}}
\label{tab:app_pt_cost}
\centering
\small
\setlength{\tabcolsep}{1.25em}
\begin{tabular}{lrccc}
    \toprule
    \multirow{2.5}{*}{\textbf{Encoder}} & \multirow{2.5}{*}{\textbf{Parameters}} & \multicolumn{3}{c}{\textbf{GPU}} \\
    \cmidrule(lr){3-5}
     & & \textbf{Count} & \textbf{Hours\textsuperscript{\dag}} & \textbf{Type} \\
    \midrule
    \texttt{OTISv2-7.1M}   & $7.1\,$M  & $2$ & ${\sim} 61$ & NVIDIA A$100$-$80$GB \\
    \texttt{OTISv2-14.8M}\textsuperscript{\ddag} & $14.8\,$M & $1$ & ${\sim} 12$ & NVIDIA A$100$-$80$GB \\
    \texttt{OTISv2-40M}\textsuperscript{\ddag}   & $40\,$M   & $1$ & ${\sim} 18$ & NVIDIA A$100$-$80$GB \\
    \texttt{OTISv2-116M}\textsuperscript{\ddag}  & $116\,$M  & $1$ & ${\sim} 27$ & NVIDIA A$100$-$80$GB \\
    \bottomrule
    \multicolumn{5}{l}{\scriptsize{\textsuperscript{\dag} GPU hours, summed over all GPUs.}} \\
    \multicolumn{5}{l}{\scriptsize{\textsuperscript{\ddag} Trained only in the scaling study (Sec.~\ref{sec:scaling}); cost summed over its eight runs.}}
\end{tabular}
\end{table}

\begin{table}[!ht]
\caption{\textbf{Pre-training hyperparameters of \texttt{OTISv2-7.1M}.}}
\label{tab:app_pt_hparams}
\centering
\small
\setlength{\tabcolsep}{0.6em}
\begin{tabular}{ll}
    \toprule
    \textbf{Hyperparameter} & \textbf{Value} \\
    \midrule
    Optimiser                                         & AdamW \citep{Loshchilov2017}, $\beta\!=\!(0.9, 0.95)$ \\
    Base learning rate (peak $=$ base $\times\, B_\text{eff}/32$)  & $1{\times}10^{-4}$ \\
    Peak / minimum learning rate                      & $1.2{\times}10^{-3}$ / $1.2{\times}10^{-4}$ \\
    Learning rate schedule                            & warmup-stable-decay, $10\,\%$ warm-up, $10\,\%$ cosine decay \\
    Layer-wise learning rate decay                    & $0.975$ \\
    Epochs                                            & $50$ \\
    Batch size (per GPU / effective)                  & $192$ / $384$ \\
    Weight decay                                      & $0.025$ \\
    Drop path                                         & $0.1$ \\
    Gradient clipping                                 & $3$ \\
    \midrule
    Patch width $p$ / context $\overline{T}$          & $24$ / $1{,}008$ \\
    Register tokens $R$                               & $4$ \\
    \midrule
    Post-fix masking probability $\pi$                & $0.33$ \\
    Masking ratio of student $\rho_s$ (random / post-fix) & $0.7$ / $0.5$, each $+\,\mathcal{U}(0, 0.15)$ per batch \\
    Masking ratio of teacher $\rho_t$                 & $0.5$ \\
    NCC weight $\lambda_\text{NCC}$                   & $0.1$ \\
    \midrule
    Self-distillation weight $\lambda_\text{SD}$ (on $\mathcal{L}_{\text{CLS}}\!+\!\mathcal{L}_{\text{patch}}$) & $0.1$ \\
    Projection heads (\texttt{[CLS]} and patch)       & $3$-layer MLP, hidden size $2{,}048$ \\
    Bottleneck size / head outputs $K$                & $256$ / $8{,}192$ \\
    Student temperature $\tau_s$                      & $0.1$ \\
    Teacher temperature $\tau_t$                      & $0.04\!\to\!0.055$ over the first $4\,\%$ of training \\
    Centring momentum                                 & $0.9$ \\
    EMA teacher momentum $m$                          & $0.996\!\to\!1$, cosine schedule \\
    Last head layer frozen                            & first $2\,\%$ of training \\
    \midrule
    KoLeo weight $\lambda_\text{KoLeo}$ (on \texttt{[CLS]}) & $0.025$ \\
    KoLeo group size                                  & $32$ \\
    \bottomrule
\end{tabular}
\end{table}

\section{Benchmark Details}
\label{sec:app_benchmark_details}
We evaluate on $143$ classification datasets from two public archives and use their official splits.
The UEA archive \citep{Bagnall2018} contributes $23$ datasets: $22$ multi-variate datasets with $2$ to $144$ variates and $26$ to $1{,}751$ time points, and Phoneme, which we use in its uni-variate UCR version ($1{,}024$ time points).
Phoneme therefore appears in both archives, so the $143$ datasets comprise $142$ distinct ones.
The UEA datasets cover speech and audio (JapaneseVowels, SpokenArabicDigits, Phoneme, Heartbeat), motion and gesture (NATOPS, RacketSports, BasicMotions, Cricket, Libras, UWaveGestureLibrary, ArticularyWordRecognition, Epilepsy), EEG and MEG (FaceDetection, FingerMovements, HandMovementDirection, SelfRegulationSCP1, SelfRegulationSCP2), ECG (AtrialFibrillation), handwriting (Handwriting, CharacterTrajectories), and other sensors (ERing, EthanolConcentration, LSST).
The UCR archive \citep{Dau2019} contributes $120$ uni-variate datasets with $15$ to $2{,}844$ time points and $2$ to $60$ classes, covering image outlines, sensors, ECG, motion, simulations, traffic, and spectroscopy.
Of the eleven variable-length UCR datasets, we use the equal-length versions of the three GestureMidAir datasets and exclude the other eight, whose equal-length versions pad the training and test splits to different lengths.
We linearly interpolate the missing values of four datasets in time.
Time series shorter than the patch width $p\!=\!24$ are right-padded with zeros.
As axial RoPE encodes only relative positions, longer time series are embedded at native length, up to their last full patch.

\section{Baseline Details}
\label{sec:app_baseline_details}
Table~\ref{tab:app_baseline_summary} summarises the baselines.
All baselines are loaded from their public checkpoints and evaluated with the frozen-encoder protocols of Appendix~\ref{sec:app_exp_details}, without fine-tuning.

\begin{table}[!ht]
\caption{\textbf{General-purpose time series encoders.} Pre-training strategy (MDM, masked data modelling; SD, self-distillation), encoder width, parameter count, and pre-training corpus size.}
\label{tab:app_baseline_summary}
\centering
\small
\setlength{\tabcolsep}{0.4em}
\begin{tabular}{llcrr}
    \toprule
    \textbf{Encoder} & \textbf{Strategy} & \textbf{Encoder width $d$} & \textbf{Parameters} & \textbf{Pre-training time points} \\
    \midrule
    \texttt{UniTS-0.4M}  \citep{Gao2024}     & MDM + task prompts & $32$   & $0.4\,$M   & $0.13\,$B$^{\dagger}$ \\
    \texttt{UniTS-1.1M}  \citep{Gao2024}     & Supervised multi-task$^{\ddagger}$ & $64$   & $1.1\,$M   & $0.13\,$B$^{\dagger}$ \\
    \texttt{UniTS-3.5M}  \citep{Gao2024}     & MDM + task prompts & $128$  & $3.5\,$M   & $0.13\,$B$^{\dagger}$ \\
    \midrule
    \texttt{MOMENT-19M}  \citep{Goswami2024} & MDM                & $512$  & $19\,$M    & $1.2\,$B \\
    \texttt{MOMENT-85M}  \citep{Goswami2024} & MDM                & $768$  & $85\,$M    & $1.2\,$B \\
    \texttt{MOMENT-308M} \citep{Goswami2024} & MDM                & $1{,}024$ & $308\,$M   & $1.2\,$B \\
    \midrule
    \texttt{OTIS-7.1M}   \citep{Turgut2026}        & MDM   & $192$  & $7.1\,$M   & $42\,$B \\
    \texttt{OTIS-43M}    \citep{Turgut2026}        & MDM   & $384$  & $43\,$M    & $42\,$B \\
    \texttt{OTIS-128M}   \citep{Turgut2026}        & MDM   & $576$  & $128\,$M   & $42\,$B \\
    \midrule
    \rowcolor{highlight}
    \texttt{OTISv2-7.1M} (ours)                    & MDM + SD + KoLeo  & $192$  & $7.1\,$M   & $45\,$B \\
    \bottomrule
    \multicolumn{5}{l}{\footnotesize $^{\dagger}$Not reported by the original work and reconstructed from its released pre-training configuration.}\\
    \multicolumn{5}{l}{\footnotesize $^{\ddagger}$The release at $d\!=\!64$ is trained fully supervised and has no masked-reconstruction head.}\\
\end{tabular}
\end{table}

\paragraph{Corpora and training tokens.}
\texttt{OTIS} and \texttt{OTISv2} are pre-trained on different corpora that share nine sources from seven domains (App.~\ref{sec:app_ts_corpus}).
\texttt{OTIS} uses $612{,}394$ samples ($41.7\,$B time points).
\texttt{OTISv2} adds the four GIFT-Eval \citep{Aksu2024} domains and the samples that \texttt{OTIS} held out for validation, for a total of $756{,}267$ samples ($45.2\,$B time points).
With a context of $\overline{T}\!=\!1{,}008$ time points and patch width $p\!=\!24$, one epoch on the two corpora yields $206.6\,$M and $219.0\,$M patch tokens, respectively.
The gap in training tokens $D$ in Table~\ref{tab:main} thus comes almost entirely from repetition.
The released \texttt{OTIS-7.1M} weights correspond to epoch $198$ ($D\!=\!40.9\,$B), and those of \texttt{OTIS-43M} and \texttt{OTIS-128M} to epochs $97$ and $99$ ($D\!=\!20.0$ and $20.5\,$B), compared with $50$ epochs for \texttt{OTISv2} ($D\!=\!10.9\,$B).
The scaling study of Section~\ref{sec:scaling} derives its data budgets from $221\,$M tokens per epoch, the count obtained if every variate filled the full context, whereas shorter samples yield fewer patches.
\texttt{MOMENT} is pre-trained for two epochs on its $1.2\,$B time points at patch width $p\!=\!8$, which gives $D\!=\!308\,$M.
\texttt{UniTS} reports no corpus size, so we reconstruct it from its released pre-training configuration.
Its $38$ task configurations draw on $25$ sources with $0.13\,$B time points in total.
Each forecasting source is sampled with sliding windows that advance by one time step, at four context lengths and for ten epochs, so each of its time points is revisited about $10^3$ times.
This gives $D\!=\!18.1\,$B patch tokens at patch width $p\!=\!16$, so \texttt{UniTS} repeats its small corpus far more often than any other encoder.
The supervised \texttt{UniTS-1.1M} trains on the same $38$ tasks for five epochs instead of ten, which gives $D\!=\!9.1\,$B.

\section{Evaluation Protocols}
\label{sec:app_exp_details}
We follow the frozen-encoder protocols of \texttt{OTIS} \citep{Turgut2026} with three changes: $k\!=\!5$ instead of $20$ neighbours, three instead of five $5$-shot episodes, and standardised instead of $\ell_2$-normalised features for the linear probe, and apply them alike to every encoder.
All metrics in Tables~\ref{tab:main} and~\ref{tab:cumulative}, and the similarity and RankMe of Table~\ref{tab:dose_response} and Figure~\ref{fig:layer_kshot}, are computed on frozen checkpoints, without masking or augmentation.

\paragraph{Feature quality.}
We draw a fixed set of $25{,}600$ samples, and every encoder within a table uses the same samples.
Tables~\ref{tab:main} and~\ref{tab:dose_response} draw from the pre-training corpus (App.~\ref{sec:app_ts_corpus}), and Table~\ref{tab:cumulative} and Figure~\ref{fig:layer_kshot} from the $101{,}858$-sample subset on which their runs were pre-trained.
Each domain contributes an equal share of these samples.
Each sample keeps its native variate count and at most its first $1{,}008$ time points, so no input is padded.
\texttt{OTIS} receives freshly initialised variate embeddings for the four domains on which it was not pre-trained.
\texttt{MOMENT} processes each sample in consecutive windows of $512$ time points.
\textit{RankMe} \citep{Garrido2023} is computed on the $25{,}600\times d$ matrix of sample embeddings, each the mean of its output patch tokens.
\textit{Patch-token cosine similarity} is the mean off-diagonal cosine similarity between the $\ell_2$-normalised output patch tokens of one sample, averaged over samples with equal weight.
Both exclude special tokens such as \texttt{[CLS]}, \texttt{[reg]}, and prompt tokens.
Over three independent draws, similarity varies by at most $0.001$ and RankMe by at most $0.9$, so larger differences between encoders are not caused by the choice of samples.
Token similarity and RankMe recorded during training (Figs.~\ref{fig:teaser} and~\ref{fig:efficiency}) are computed differently.
They use masked training batches, pool similarity over all token pairs of a batch, and compute RankMe on the patch tokens of a batch rather than on sample embeddings.
Training batches pad samples to a common variate count, and the padded tokens are either excluded or corrected by a constant offset per run, measured at its final checkpoint.
These values are therefore compared only within their own figure.

\paragraph{Embeddings.}
We use the \texttt{[CLS]} token for \texttt{OTISv2}, as self-distillation supervises it, and the mean of the patch tokens for all other encoders.
\texttt{MOMENT} splits series longer than $512$ time points into consecutive windows of $512$ time points and averages the valid patch tokens of all windows; shorter series are right-padded and masked.
\texttt{OTIS} interpolates its time embedding beyond its trained context of $1{,}008$ time points and receives freshly initialised variate embeddings for every unseen dataset.
\texttt{UniTS} has no prompt and task tokens for unseen datasets, so these are freshly initialised.

\paragraph{$5$-shot prototypical classification.}
This protocol tests how well a few labelled samples describe each class.
In each episode, every class contributes $k\!=\!5$ support samples drawn from the training split without replacement, or all of its samples if it has fewer.
The mean of the support embeddings of a class forms its prototype.
Prototypes and test embeddings are $\ell_2$-normalised, and each test sample takes the label of the prototype with the highest cosine similarity.
We average the balanced accuracy over three episodes with new support sets.
A fixed seed gives every encoder the same support sets.

\paragraph{$k$-NN classification.}
This protocol tests whether samples of the same class lie close together in the embedding space, using the full training split and no fitted parameters.
Following \texttt{DINO} \citep{Caron2021}, train and test embeddings are $\ell_2$-normalised, and each test sample retrieves its $k\!=\!5$ nearest training samples by cosine similarity, with $k$ capped at the size of the training split.
Neighbour $i$ votes for its class with weight $w_i\!=\!\exp(\cos_i/\tau)$ and $\tau\!=\!0.07$, and the class that collects the largest total weight is predicted.
We keep $k$ small, as some datasets have fewer than five training samples in a class and a larger neighbourhood would mix classes.

\paragraph{Linear probing.}
This protocol tests whether the classes are linearly separable in the embedding space.
Embeddings are standardised with training-split statistics.
We fit an $\ell_2$-regularised multinomial logistic regression with L-BFGS.
The regularisation strength is chosen from $C\!\in\!\{10^{-2}, 10^{-1}, 1, 10, 10^{2}\}$ by balanced accuracy on a stratified $20\,\%$ hold-out split of the training data.
Datasets with fewer than $50$ training samples or fewer than five per class use stratified cross-validation instead.
The final probe is then trained on the full training split.

\section{Per-Block Representation Quality}
\label{sec:app_layer_kshot}

\begin{figure}[!t]
    \centering
    \includegraphics[width=1.0\linewidth]{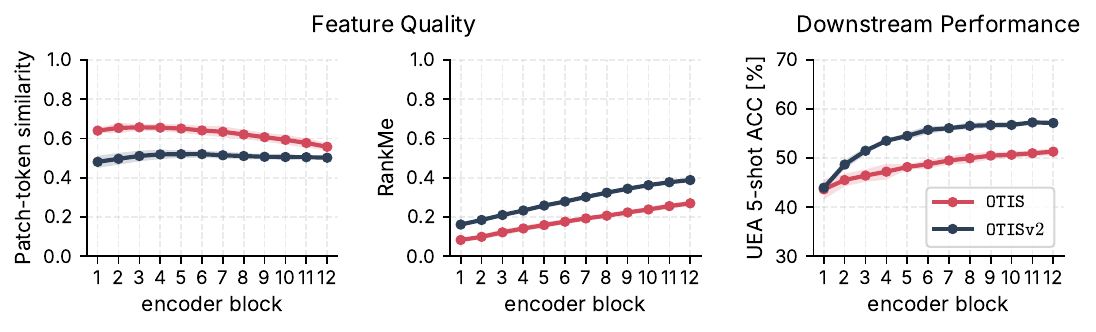}
    \caption{\textbf{The advantage of \texttt{OTISv2} builds up over depth.}
    Token similarity, RankMe, and $5$-shot accuracy at every block of the matched runs of \texttt{OTIS} and \texttt{OTISv2} ($N\!=\!7.1\,$M, $D\!=\!1.60\,$B, $D_C\!=\!32\,$M, $12$ blocks, $d\!=\!192$).
    Lines show the mean over three pre-training seeds, and shading one standard deviation.
    Similarity and RankMe follow the protocol of Table~\ref{tab:cumulative} (App.~\ref{sec:app_exp_details}), and at block $12$ they equal its first and last rows.
    \textbf{(Left)}~Patch-token cosine similarity.
    \textbf{(Centre)}~RankMe normalised by $d$.
    \textbf{(Right)}~UEA $5$-shot balanced accuracy on the \texttt{[CLS]} token of each block.}
    \label{fig:layer_kshot}
\end{figure}

\paragraph{Setup.}
Table~\ref{tab:cumulative} measures the encoders only at their output, so it leaves open whether the advantage of \texttt{OTISv2} forms in the last blocks or builds up over depth.
If it formed only at the output, the two encoders would differ only in the last blocks.
Figure~\ref{fig:layer_kshot} therefore probes the matched $7.1\,$M runs of \texttt{OTIS} and \texttt{OTISv2} (the first and last rows of Tab.~\ref{tab:cumulative}, three pre-training seeds each) at the output of every block, which we pass through the final normalisation layer of the encoder.
Token similarity and RankMe follow Appendix~\ref{sec:app_exp_details} on the samples of Table~\ref{tab:cumulative}, so block $12$ reproduces the values of that table.
$5$-shot accuracy uses the \texttt{[CLS]} token of each block on the $23$ UEA datasets.

\paragraph{Results.}
In both encoders, RankMe rises with depth.
Normalised by the encoder width $d$, the RankMe of \texttt{OTISv2} is higher at every block, rising from $0.16$ at block $1$ to $0.39$ at block $12$, compared with $0.08$ to $0.27$ for \texttt{OTIS}.
\texttt{OTISv2} also has lower token similarity at every block, between $0.48$ and $0.52$, whereas that of \texttt{OTIS} stays between $0.64$ and $0.66$ up to block $6$ and then falls to $0.56$ at block $12$.
Both orderings hold at every block in each of the three seeds, so \texttt{OTISv2} mitigates feature collapse throughout the encoder, not only at its output.
$5$-shot accuracy also rises with depth in both encoders and levels off over the last blocks of \texttt{OTISv2} ($57.3\,\%$ and $57.2\,\%$ at blocks $11$ and $12$).
At block $1$, both encoders score almost the same ($44.0\,\%$ for \texttt{OTISv2} against $43.6\,\%$ for \texttt{OTIS}).
From block $2$ onwards, \texttt{OTISv2} leads in every seed, and its margin grows from $3.1\,$pp at block $2$ to $6.3\,$pp at block $4$, then stays between $5.8$ and $7.0\,$pp.
The advantage of \texttt{OTISv2} thus builds up over the first blocks and is not limited to the last block.

\section{Patch-Token Norms}
\label{sec:app_token_norms}

\begin{figure}[!ht]
    \centering
    \includegraphics[width=\linewidth]{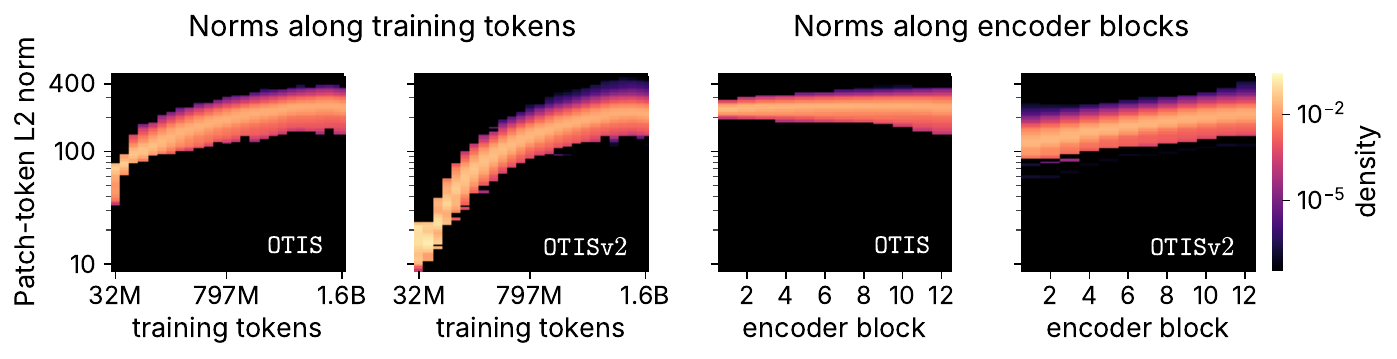}
    \caption{\textbf{Neither encoder develops high-norm artefact tokens during pre-training.}
    Patch-token $\ell_2$ norms of the matched runs of \texttt{OTIS} and \texttt{OTISv2} ($N\!=\!7.1\,$M, $D\!=\!1.60\,$B, $D_C\!=\!32\,$M), one pre-training seed each.
    Colour gives the density of tokens.
    \textbf{(Left two)}~Last encoder block over training.
    \textbf{(Right two)}~All encoder blocks at the end of pre-training.}
    \label{fig:tokennorms}
\end{figure}

\paragraph{Setup.}
Large Vision Transformers \citep{Dosovitskiy2021} develop high-norm artefact tokens that store global information \citep{Darcet2024}.
If such tokens appeared in our encoders, a separate population of patch tokens with much larger norms would emerge.
To test this, Figure~\ref{fig:tokennorms} records the $\ell_2$ norms of the patch tokens at the output of every block, before the final normalisation layer, for the matched $7.1\,$M runs of \texttt{OTIS} and \texttt{OTISv2} (one pre-training seed each) on $4{,}096$ pre-training samples with all patches visible.

\paragraph{Results.}
Neither encoder develops a separate high-norm population, so artefact tokens do not explain the higher token similarity of \texttt{OTIS}.
The encoders differ only in how their norms grow.
At the last block, the norms at the start of training lie between $30$ and $100$ for \texttt{OTIS} and between $10$ and $20$ for \texttt{OTISv2}.
After training, the norms of \texttt{OTIS} already lie in a narrow band of $200$ to $300$ at the first block, whereas those of \texttt{OTISv2} start at about half these values and grow linearly with depth.
Both end in a band of $150$ to $300$ at the last block.
\texttt{OTISv2} thus defers norm growth to later blocks rather than suppressing it.
As the two runs differ in architecture, training stabilisers, and objective, this comparison does not show which change causes the difference in norm growth.

\section{Single-Sample Encoder Inspection on 12-Lead ECG}
\label{sec:app_ecg_inspection}
To illustrate how the two encoders represent a single sample, we inspect the last block of the frozen \texttt{OTIS-7.1M} and \texttt{OTISv2-7.1M} encoders of Table~\ref{tab:main} on four $12$-lead ECGs from the pre-training corpus.
Each sample is embedded without masking as a grid of $12$ leads by $42$ time patches, $504$ patches in total.
Table~\ref{tab:app_ecg_summary} summarises the norm of each patch token after the final normalisation layer, a joint PCA of the patch tokens, and the attention that each patch receives.

\begin{table}[!ht]
\caption{\textbf{Last-block properties on four $12$-lead ECG samples} ($504$ patches per sample).
Norms are taken after the final normalisation layer, and attention is averaged over the patch queries of each head.
Patch energy is the standard deviation of the raw signal in a patch, and the effective keys of a query are the exponential of the entropy of its attention.
The two checkpoints are not matched, so these contrasts are illustrative only (App.~\ref{sec:app_ecg_inspection}).}
\label{tab:app_ecg_summary}
\centering
\small
\setlength{\tabcolsep}{1.0em}
\begin{tabular}{lcc}
    \toprule
                                                                        & \texttt{OTIS}             & \texttt{OTISv2}           \\
    \midrule
    Token $\ell_2$ norm range                                              & $[6.65,\,10.34]$          & $[11.50,\,20.47]$         \\
    Token $\ell_2$ norm spread (max\,/\,min)                               & $1.56\times$              & $1.78\times$              \\
    Spearman $\rho$(token norm, raw patch energy)                       & $-0.18$                   & $+0.39$                   \\
    \midrule
    $\max|\rho|$(PC1--PC3, raw patch energy)                            & $0.62$ (PC2)              & $0.45$ (PC2)              \\
    PCA explained variance, top-3 sum                                   & $0.35$                    & $0.38$                    \\
    \midrule
    Attention max received (uniform $= 0.00198$)                        & $0.0132$ ($6.7\times$)    & $0.0187$ ($9.4\times$)    \\
    Effective keys per patch query                                      & $321$                     & $220$                     \\
    Correlation between patch-query maps (mean)                         & $0.72$                    & $0.09$                    \\
    \bottomrule
\end{tabular}
\end{table}

\paragraph{Caveat.}
The two checkpoints are not matched.
\texttt{OTIS} was pre-trained on its own corpus (App.~\ref{sec:app_baseline_details}) with $38$ variate embeddings and no registers, and \texttt{OTISv2} on the corpus of Appendix~\ref{sec:app_ts_corpus} with four registers.
As the corpus, variate layout, and training length differ in addition to the recipe, this section only illustrates single samples and supports no claim of the paper.

\begin{figure}[!ht]
    \centering
    \begin{subfigure}[t]{\linewidth}
        \includegraphics[width=\linewidth]{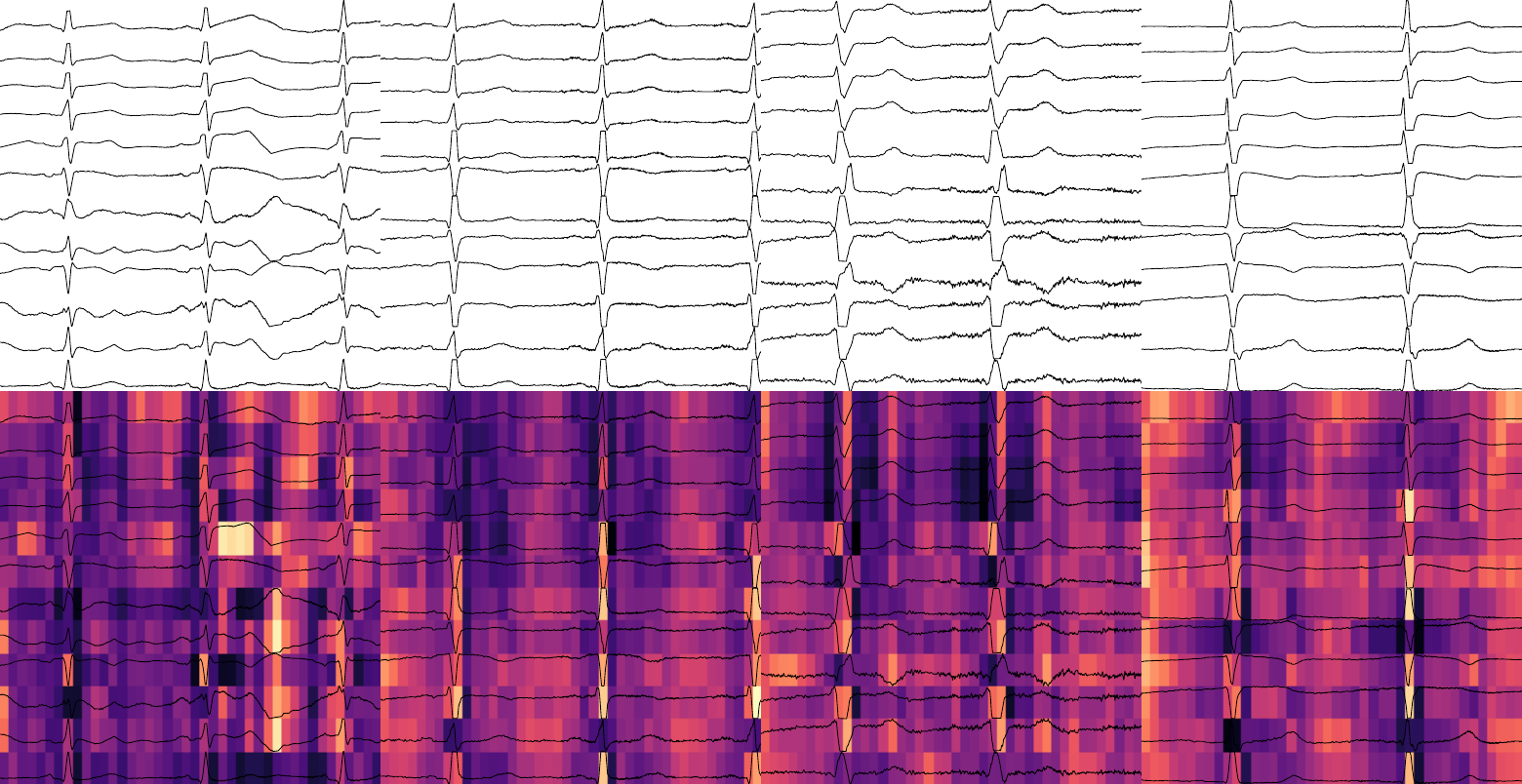}
        \caption{\texttt{OTIS}.}
        \label{fig:app_ecg_norms:v1}
    \end{subfigure}\\[0.4em]
    \begin{subfigure}[t]{\linewidth}
        \includegraphics[width=\linewidth]{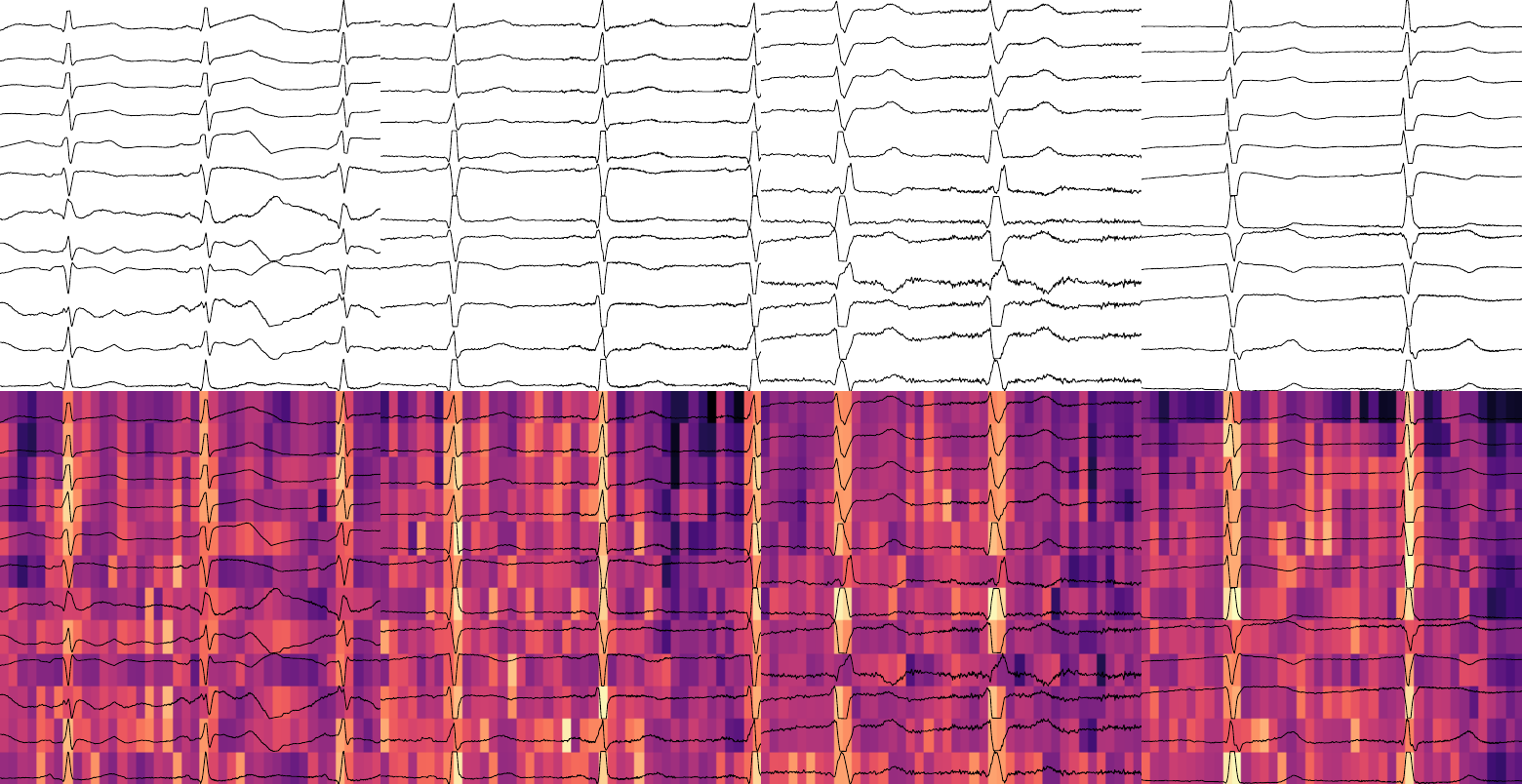}
        \caption{\texttt{OTISv2}.}
        \label{fig:app_ecg_norms:v2}
    \end{subfigure}
    \caption{\textbf{Both encoders keep token norms in a narrow band without outliers.}
    Per-patch token $\ell_2$ norms after the final normalisation layer on four $12$-lead ECGs.
    In each subfigure, the top row shows the raw signal, and the bottom row shows the norm of each patch with the signal overlaid, on a colour scale per encoder shared across the four samples.}
    \label{fig:app_ecg_norms}
\end{figure}

\paragraph{Token norms.}
After the final normalisation layer, the token norms of \texttt{OTIS} span $6.65$ to $10.34$ and those of \texttt{OTISv2} span $11.50$ to $20.47$ (Fig.~\ref{fig:app_ecg_norms}), so the largest norm exceeds the smallest by only $1.56\times$ and $1.78\times$.
Neither encoder thus produces outlier tokens, in line with Appendix~\ref{sec:app_token_norms}, whose norms are taken before this layer.
The norm correlates only weakly with patch energy, the standard deviation of its raw signal, negatively for \texttt{OTIS} (Spearman $\rho\!=\!-0.18$) and positively for \texttt{OTISv2} ($\rho\!=\!0.39$).

\begin{figure}[!ht]
    \centering
    \includegraphics[width=\linewidth]{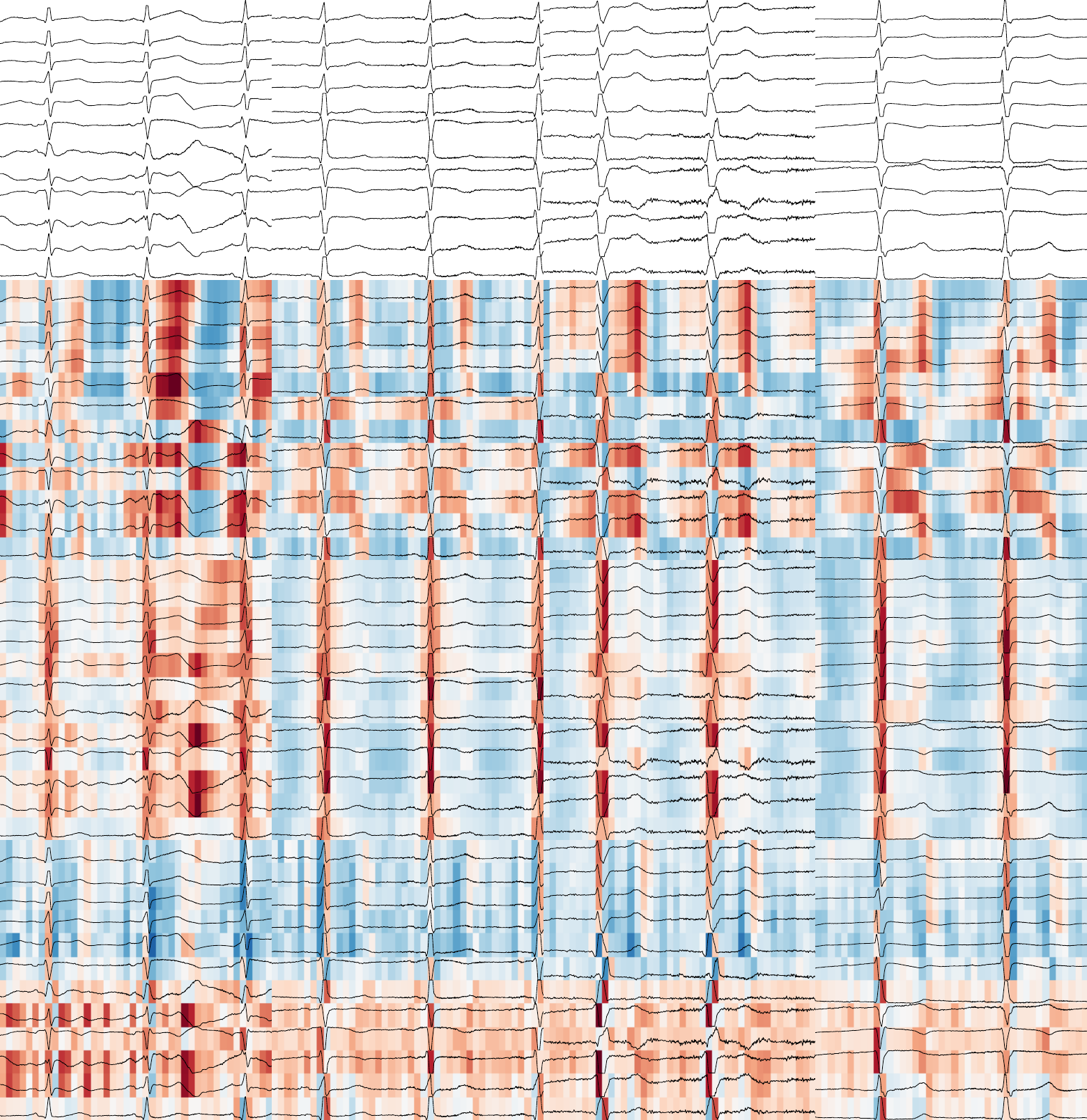}
    \caption{\textbf{The leading patch-token components of \texttt{OTIS} vary smoothly across neighbouring patches.}
    Joint PCA of the last-block patch tokens of \texttt{OTIS} over the four samples.
    Row $1$ shows the raw signal, and rows $2$ to $4$ show PC1 to PC3 with the signal overlaid, on a symmetric colour scale shared across samples.}
    \label{fig:app_ecg_pca:v1}
\end{figure}

\begin{figure}[!ht]
    \centering
    \includegraphics[width=\linewidth]{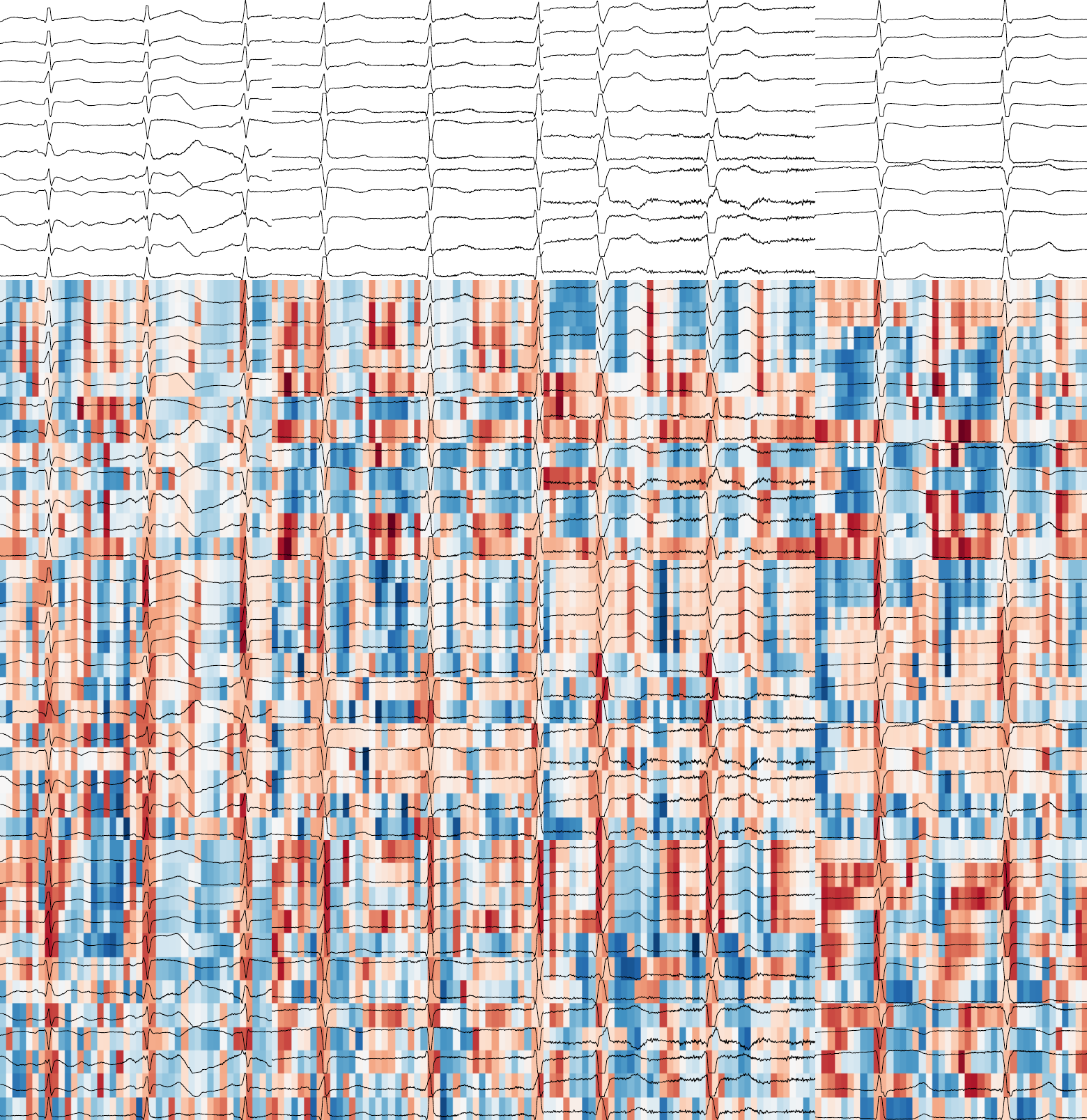}
    \caption{\textbf{The leading patch-token components of \texttt{OTISv2} vary more from patch to patch.}
    Joint PCA of the last-block patch tokens of \texttt{OTISv2}, with the same layout as Figure~\ref{fig:app_ecg_pca:v1}.}
    \label{fig:app_ecg_pca:v2}
\end{figure}

\paragraph{PCA.}
The leading components of \texttt{OTIS} vary smoothly across neighbouring patches (Fig.~\ref{fig:app_ecg_pca:v1}).
One is aligned with the QRS complexes, the large deflections of each heartbeat, and correlates with patch energy ($\rho\!=\!0.62$), and another varies mainly between leads.
Those of \texttt{OTISv2} vary more from patch to patch, with weaker structure across leads and time (Fig.~\ref{fig:app_ecg_pca:v2}).
On these four samples, the top three components explain a similar share of the variance in \texttt{OTIS} and \texttt{OTISv2} ($35\,\%$ and $38\,\%$).
Neighbouring patch tokens of \texttt{OTISv2} thus differ more, consistent with its lower token similarity.

\begin{figure}[!ht]
    \centering
    \includegraphics[width=\linewidth]{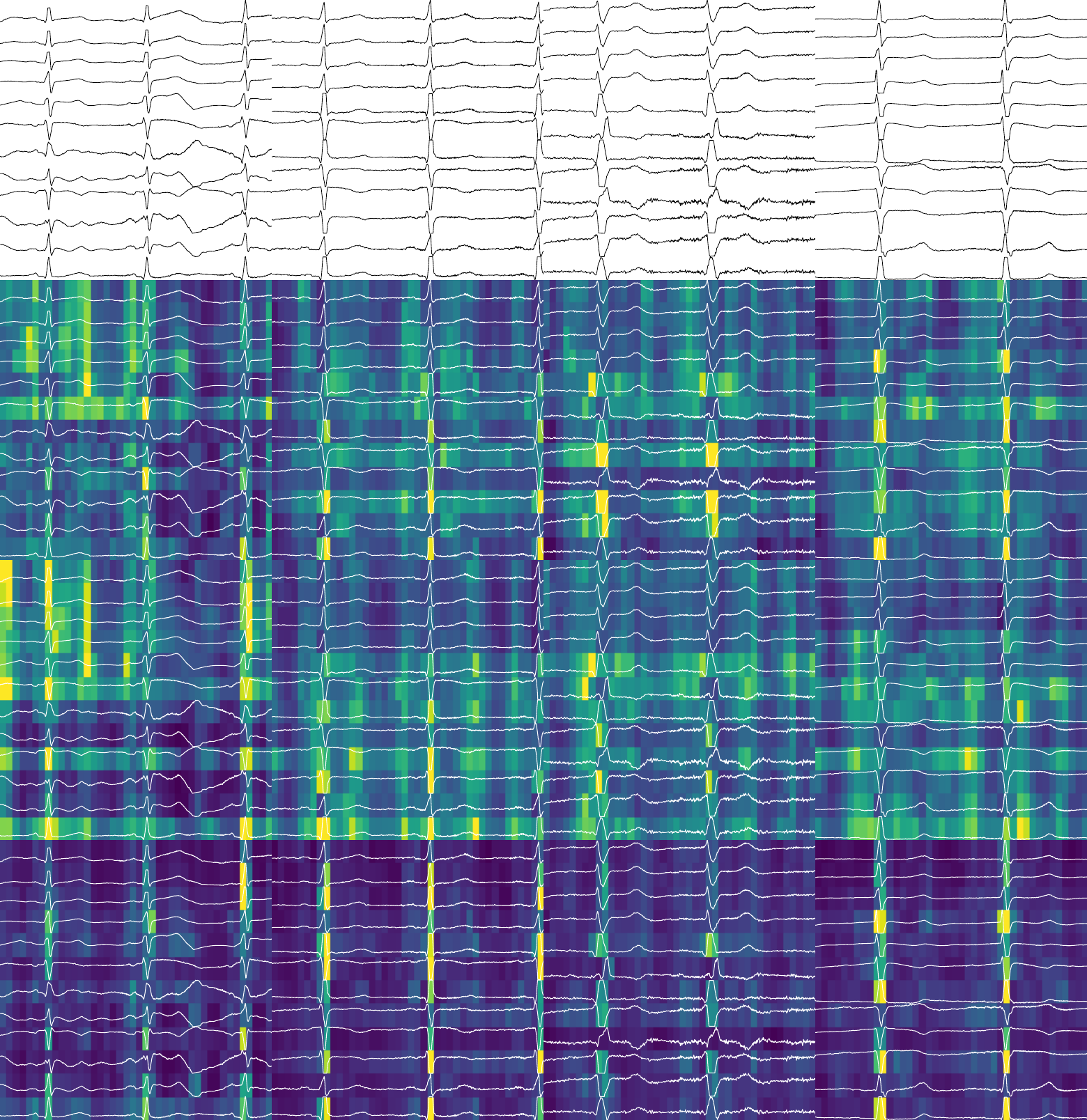}
    \caption{\textbf{\texttt{OTIS} attends to QRS-aligned patches in one head.}
    Mean last-block attention that each patch receives from all query patches, per head, with the signal overlaid.
    Row $1$ shows the raw signal, and rows $2$ to $4$ show the three heads.}
    \label{fig:app_ecg_attn:v1}
\end{figure}

\begin{figure}[!ht]
    \centering
    \includegraphics[width=\linewidth]{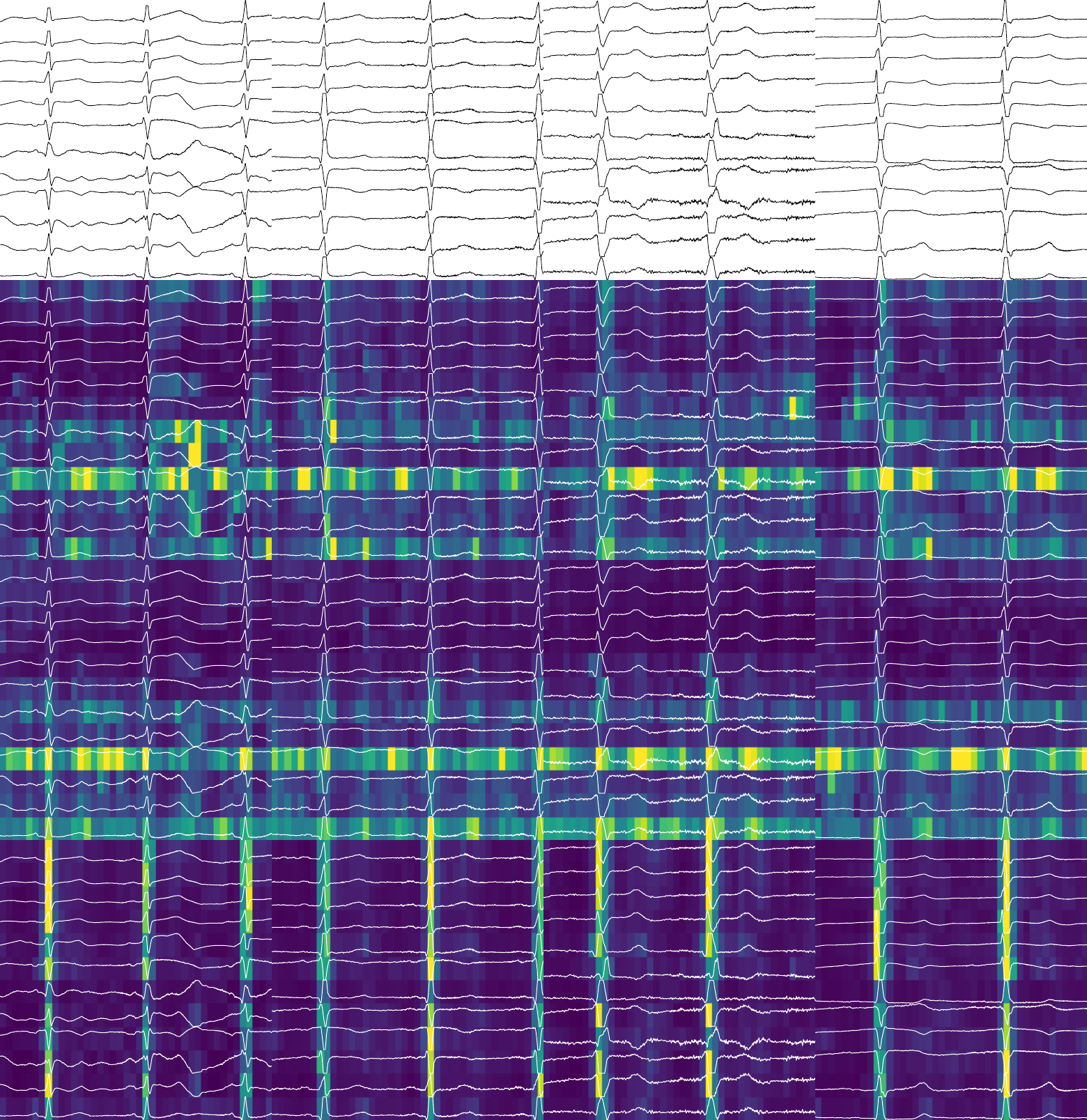}
    \caption{\textbf{\texttt{OTISv2} attends to QRS-aligned patches in one head and to a few leads in the other two.}
    Per-head last-block attention of \texttt{OTISv2}, with the same layout as Figure~\ref{fig:app_ecg_attn:v1}.}
    \label{fig:app_ecg_attn:v2}
\end{figure}

\begin{figure}[!ht]
    \centering
    \begin{subfigure}[t]{\linewidth}
        \includegraphics[width=\linewidth]{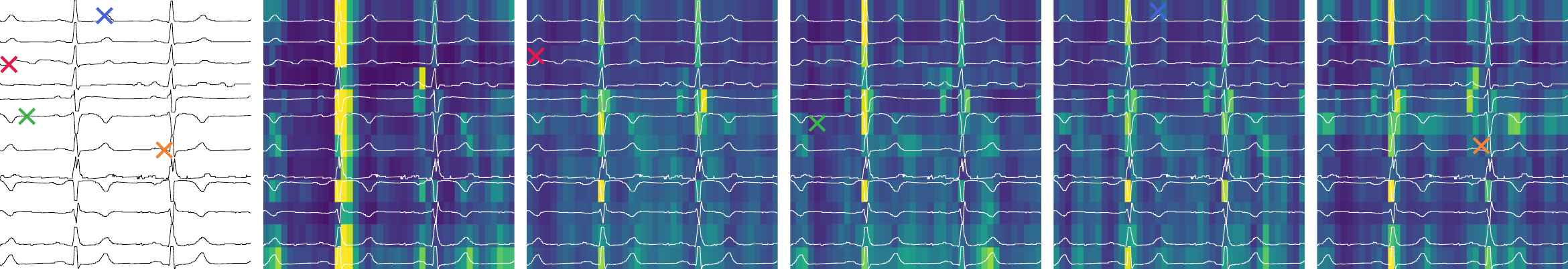}
        \caption{\texttt{OTIS}.}
        \label{fig:app_ecg_random_query:v1}
    \end{subfigure}\\[0.4em]
    \begin{subfigure}[t]{\linewidth}
        \includegraphics[width=\linewidth]{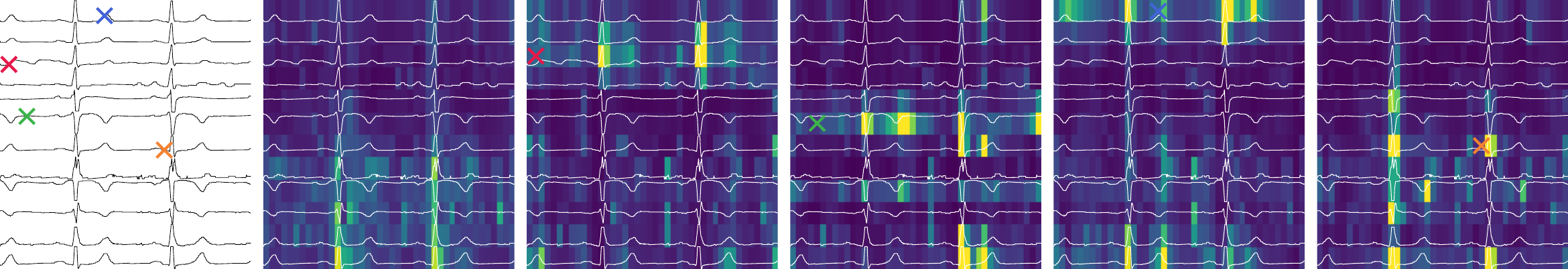}
        \caption{\texttt{OTISv2}.}
        \label{fig:app_ecg_random_query:v2}
    \end{subfigure}
    \caption{\textbf{Each \texttt{OTISv2} query attends to its own set of patches.}
    Head-averaged last-block attention of the \texttt{[CLS]} token and four random patch queries on one $12$-lead ECG.
    Column $1$ shows the signal with the query positions marked by coloured crosses, column $2$ the attention of \texttt{[CLS]}, and columns $3$ to $6$ the attention of each patch query.}
    \label{fig:app_ecg_random_query}
\end{figure}

\paragraph{Attention.}
Uniform attention would give each patch $1/504\!=\!0.00198$ of the attention.
Neither encoder attends uniformly (Figs.~\ref{fig:app_ecg_attn:v1} and~\ref{fig:app_ecg_attn:v2}).
In both encoders, one head attends to the QRS-aligned patches, and the $10\,\%$ of patches with the highest energy receive $17\,\%$ of the attention in \texttt{OTIS} and $23\,\%$ in \texttt{OTISv2}, against $10\,\%$ under uniform attention.
In \texttt{OTISv2}, each of the other two heads concentrates on a few leads.
The most attended patch receives $6.7\times$ uniform attention in \texttt{OTIS} and $9.4\times$ in \texttt{OTISv2}.
The attention of a patch query spreads over $321$ and $220$ effective keys, respectively, measured as the exponential of its entropy, so \texttt{OTISv2} is only moderately more concentrated.
The attention maps of single queries show the clearest difference (Fig.~\ref{fig:app_ecg_random_query}).
In \texttt{OTIS}, four random patch queries produce similar maps focused on QRS-aligned patches (pairwise correlation $0.66$ to $0.86$), and the \texttt{[CLS]} token focuses on one QRS-aligned column of patches ($17\times$ uniform attention).
In \texttt{OTISv2}, each patch query attends to its own set of patches (pairwise correlation $-0.06$ to $0.23$) and puts $17$ to $24\,\%$ of its attention on its own lead, against $8\,\%$ under uniform attention, while the \texttt{[CLS]} token attends diffusely.
On these samples, the patch queries of \texttt{OTIS} thus draw on the same few patches, whereas each query of \texttt{OTISv2} gathers its own context.

\section{Encoder Configurations}
\label{sec:app_encoder_variants}
Table~\ref{tab:app_encoder_variants} lists the four configurations of the encoder $f(\cdot)$.
The released \texttt{OTISv2} has $7.1\,$M parameters, and the scaling study of Section~\ref{sec:scaling} uses all four sizes, the $14.8\,$M, $40\,$M, and $116\,$M variants only there.
All variants keep the per-head dimension at $64$ and the MLP size at four times the encoder width, and grow in both width and depth.

\begin{table}[!ht]
\caption{\textbf{Encoder configurations of \texttt{OTISv2}.}}
\label{tab:app_encoder_variants}
\centering
\normalsize
\setlength{\tabcolsep}{0.6em}
\begin{tabular}{lcccccr}
    \toprule
    \textbf{Encoder} & \textbf{Blocks} & \textbf{Encoder width \textit{d}} & \textbf{MLP size} & \textbf{Heads} & \textbf{$d_h$} & \textbf{Parameters} \\
    \midrule
    \texttt{OTISv2-7.1M}   & $12$ & $192$ &  $768$ & $3$ & $64$ &  $7.1\,$M \\
    \texttt{OTISv2-14.8M}  & $14$ & $256$ & $1{,}024$ & $4$ & $64$ & $14.8\,$M \\
    \texttt{OTISv2-40M}    & $24$ & $320$ & $1{,}280$ & $5$ & $64$ &   $40\,$M \\
    \texttt{OTISv2-116M}   & $36$ & $448$ & $1{,}792$ & $7$ & $64$ &  $116\,$M \\
    \bottomrule
\end{tabular}
\end{table}

\section{Zero-Shot Forecasting}
\label{sec:app_forecasting}

\paragraph{Setup.}
\texttt{OTISv2} is evaluated as an encoder for discriminative tasks, but its pre-training decoder can already forecast, as it reconstructs masked patches at the end of a series.
If the feature-space objectives improve the representation as a whole, this forecast should also improve over \texttt{OTIS} without any training.
Table~\ref{tab:app_forecasting} therefore compares the released \texttt{OTIS} and \texttt{OTISv2} encoders on seven standard forecasting datasets \citep{Zhou2021, Wu2021, Lai2018}, with encoder and decoder frozen, so that no parameter is trained or tuned.
Each test window gives the encoder a lookback of $336$ time points and masks the next $96$, $192$, or $336$ time points, which the decoder reconstructs; for the short weekly ILI series, the lookback is $96$ and the horizons are $24$, $48$, and $72$.
ETT, Weather, and ILI enter as multi-variate series, and Electricity as one uni-variate series per client.
Each variate is normalised with the statistics of the training split, test windows advance by one time point, and we report the mean squared error (MSE) over the horizon, averaged over the three horizons.
We do not compare \texttt{MOMENT} and \texttt{UniTS}, as the released \texttt{MOMENT} checkpoints contain no reconstruction head and \texttt{UniTS} was pre-trained on four of the seven datasets.
We exclude the common Traffic dataset, as our pre-training corpus contains $803$ of its $862$ series, including $95\,\%$ of its test period, through the GIFT-Eval collection \citep{Aksu2024}.

\begin{table}[!t]
\caption{\textbf{Without any training, \texttt{OTISv2} has the lowest average forecasting error.}
Zero-shot MSE ($\downarrow$) of the frozen released encoders and their pre-training decoders, averaged over the horizons $96$, $192$, and $336$ ($24$, $48$, and $72$ for ILI).
Bold marks the best encoder per dataset.}
\label{tab:app_forecasting}
\centering
\small
\setlength{\tabcolsep}{0.45em}
\begin{tabular}{lcccccccc}
    \toprule
    \textbf{Encoder} & ETTh1 & ETTh2 & ETTm1 & ETTm2 & Weather & Electricity & ILI & \textbf{Avg} \\
    \midrule
    \texttt{OTIS-7.1M}   & $1.916$ & $1.611$ & $2.230$ & $1.500$ & $0.565$ & $0.797$ & $7.295$ & $2.273$ \\
    \texttt{OTIS-43M}    & $1.161$ & $0.479$ & $1.409$ & $0.479$ & $0.394$ & $0.823$ & $5.414$ & $1.451$ \\
    \texttt{OTIS-128M}   & $1.071$ & $\mathbf{0.317}$ & $\mathbf{1.068}$ & $\mathbf{0.381}$ & $0.383$ & $0.866$ & $5.868$ & $1.422$ \\
    \midrule
    \rowcolor{highlight}
    \texttt{OTISv2-7.1M} & $\mathbf{1.048}$ & $0.427$ & $1.506$ & $0.546$ & $\mathbf{0.257}$ & $\mathbf{0.617}$ & $\mathbf{5.074}$ & $\mathbf{1.354}$ \\
    \bottomrule
\end{tabular}
\end{table}

\paragraph{Results.}
\texttt{OTISv2} forecasts better than \texttt{OTIS} of the same size on all seven datasets, and its average MSE of $1.354$ is the lowest, below the $1.422$ of the $18\times$ larger \texttt{OTIS-128M}.
\texttt{OTISv2} also has the lowest mean absolute error ($0.719$ against $0.761$).
It forecasts best on ETTh1, Weather, Electricity, and ILI, whereas \texttt{OTIS-128M} forecasts best on ETTh2, ETTm1, and ETTm2.
The gains of the feature-space objectives thus transfer to a generative task without any training, although thorough forecasting evaluation remains future work.

\section{Robustness to Input Corruptions}
\label{sec:app_corruption}

\paragraph{Setup.}
The benchmarks of Table~\ref{tab:main} contain clean recordings, whereas real sensors add noise, spikes, gaps, and failures.
If the frozen embeddings were robust, accuracy under mild corruptions would stay close to its clean value.
Table~\ref{tab:app_corruption} therefore corrupts the test splits of the $23$ UEA datasets, keeps the training splits clean, and evaluates the encoders of Table~\ref{tab:main} with the three protocols of Appendix~\ref{sec:app_exp_details}.
We apply four corruptions at a mild severity: Gaussian noise with $5\,\%$ of the standard deviation of each variate, spikes of $\pm5$ standard deviations on $0.5\,\%$ of the time points, three missing segments that cover $10\,\%$ of one variate per sample and are filled with the mean of its remaining values, and variate dropout, which removes one variate per dataset.
All encoders receive the same corrupted inputs, which are not re-normalised.
Besides the \texttt{[CLS]} embedding of \texttt{OTISv2}, we also evaluate its mean-pooled patch tokens, the embedding used for all other encoders.

\paragraph{Results.}
Under every corruption and protocol, \texttt{OTISv2} loses more accuracy than each \texttt{OTIS} encoder.
Its UEA $5$-shot accuracy drops by $1.6$ to $5.7\,$pp, against $0.1$ to $4.0\,$pp for \texttt{OTIS-128M}, with the largest gap under noise ($3.18$ against $0.07\,$pp).
\texttt{OTISv2} still matches \texttt{OTIS-128M} under spikes ($56.98\,\%$ against $56.97\,\%$) and leads under missing segments ($58.39\,\%$ against $58.19\,\%$), but trails it by $1.81\,$pp under noise and $0.58\,$pp under variate dropout.
Under $k$-NN classification and linear probing, \texttt{OTIS-128M}, \texttt{OTIS-43M}, or \texttt{MOMENT-308M} leads every corrupted setting, by $0.75$ to $3.84\,$pp over \texttt{OTISv2}.
Most of this sensitivity enters through the \texttt{[CLS]} embedding, as the mean-pooled patch tokens of \texttt{OTISv2} lose less accuracy in $11$ of the $12$ settings, for example $1.10$ instead of $3.18\,$pp under noise in $5$-shot classification.
The encoder itself is also more sensitive to noise, as even its mean-pooled patch tokens lose more than those of \texttt{OTIS-7.1M} ($1.10$ against $0.04\,$pp).
Three of the $23$ datasets, Handwriting, CharacterTrajectories, and SpokenArabicDigits, account for $85\,\%$ of the additional relative loss of \texttt{OTISv2} under noise in $5$-shot classification.
The selective \texttt{[CLS]} summary that yields \texttt{OTISv2} its clean advantage thus also passes input faults on more strongly, so for noisy inputs its mean-pooled patch tokens are the more robust embedding.

\begin{table}[!ht]
\caption{\textbf{\texttt{OTISv2} loses more accuracy under corruptions, mostly through its \texttt{[CLS]} embedding.}
Balanced accuracy ($\uparrow$) on the $23$ UEA datasets with clean training splits and corrupted test splits.
Bold marks the best encoder per protocol and condition.
\texttt{UniTS} and the two smaller \texttt{MOMENT} encoders, which lead in no setting, are omitted, and the clean accuracies reproduce Table~\ref{tab:main} within $0.1\,$pp.}
\label{tab:app_corruption}
\centering
\small
\setlength{\tabcolsep}{1.0em}
\begin{tabular}{lccccc}
    \toprule
    \textbf{Encoder} & Clean & Noise & Spikes & Missing & Dropout\textsuperscript{\dag} \\
    \midrule
    \multicolumn{6}{l}{\textit{$5$-shot prototypical classification}} \\
    \texttt{MOMENT-308M} & $52.84$ & $52.20$ & $50.04$ & $52.83$ & $52.82$ \\
    \texttt{OTIS-7.1M}   & $55.53$ & $55.49$ & $53.43$ & $55.02$ & $53.75$ \\
    \texttt{OTIS-43M}    & $57.53$ & $57.48$ & $55.49$ & $57.28$ & $55.41$ \\
    \texttt{OTIS-128M}   & $58.70$ & $\mathbf{58.63}$ & $56.97$ & $58.19$ & $\mathbf{56.61}$ \\
    \rowcolor{highlight}
    \texttt{OTISv2-7.1M} & $\mathbf{60.00}$ & $56.82$ & $\mathbf{56.98}$ & $\mathbf{58.39}$ & $56.04$ \\
    \texttt{OTISv2-7.1M}, mean-pooled & $55.77$ & $54.67$ & $53.45$ & $55.08$ & $53.06$ \\
    \midrule
    \multicolumn{6}{l}{\textit{$k$-NN classification}} \\
    \texttt{MOMENT-308M} & $58.37$ & $57.85$ & $56.09$ & $58.04$ & $58.34$ \\
    \texttt{OTIS-7.1M}   & $64.42$ & $64.28$ & $61.36$ & $63.51$ & $59.88$ \\
    \texttt{OTIS-43M}    & $64.80$ & $65.03$ & $62.57$ & $64.55$ & $61.91$ \\
    \texttt{OTIS-128M}   & $65.78$ & $\mathbf{65.69}$ & $\mathbf{63.96}$ & $\mathbf{64.96}$ & $\mathbf{62.26}$ \\
    \rowcolor{highlight}
    \texttt{OTISv2-7.1M} & $\mathbf{66.17}$ & $61.85$ & $61.96$ & $63.39$ & $61.51$ \\
    \texttt{OTISv2-7.1M}, mean-pooled & $62.32$ & $60.86$ & $59.14$ & $61.80$ & $59.13$ \\
    \midrule
    \multicolumn{6}{l}{\textit{Linear probing}} \\
    \texttt{MOMENT-308M} & $64.37$ & $62.97$ & $58.03$ & $63.12$ & $\mathbf{64.54}$ \\
    \texttt{OTIS-7.1M}   & $67.73$ & $66.95$ & $62.60$ & $66.30$ & $61.96$ \\
    \texttt{OTIS-43M}    & $69.27$ & $\mathbf{69.41}$ & $66.41$ & $68.71$ & $63.95$ \\
    \texttt{OTIS-128M}   & $69.18$ & $69.21$ & $\mathbf{66.43}$ & $\mathbf{68.73}$ & $64.04$ \\
    \rowcolor{highlight}
    \texttt{OTISv2-7.1M} & $\mathbf{69.85}$ & $66.10$ & $64.46$ & $66.14$ & $63.35$ \\
    \texttt{OTISv2-7.1M}, mean-pooled & $68.51$ & $65.80$ & $63.07$ & $66.44$ & $62.93$ \\
    \bottomrule
    \multicolumn{6}{l}{\scriptsize \textsuperscript{\dag}On the $22$ multi-variate datasets.}
\end{tabular}
\end{table}


\clearpage
\section*{NeurIPS Paper Checklist}

\begin{enumerate}

\item {\bf Claims}
    \item[] Question: Do the main claims made in the abstract and introduction accurately reflect the paper's contributions and scope?
    \item[] Answer: \answerYes{} 
    \item[] Justification: The main claims made in the abstract and introduction accurately reflect the paper's contributions and scope.

\item {\bf Limitations}
    \item[] Question: Does the paper discuss the limitations of the work performed by the authors?
    \item[] Answer: \answerYes{} 
    \item[] Justification: The paper discusses the limitations of the work.

\item {\bf Theory assumptions and proofs}
    \item[] Question: For each theoretical result, does the paper provide the full set of assumptions and a complete (and correct) proof?
    \item[] Answer: \answerNA{} 
    \item[] Justification: This work is a fully empirical study. 

    \item {\bf Experimental result reproducibility}
    \item[] Question: Does the paper fully disclose all the information needed to reproduce the main experimental results of the paper to the extent that it affects the main claims and/or conclusions of the paper (regardless of whether the code and data are provided or not)?
    \item[] Answer: \answerYes{} 
    \item[] Justification: The paper fully discloses all the information needed to reproduce the main experimental results.

\item {\bf Open access to data and code}
    \item[] Question: Does the paper provide open access to the data and code, with sufficient instructions to faithfully reproduce the main experimental results, as described in supplemental material?
    \item[] Answer: \answerYes{} 
    \item[] Justification: The paper provides open access to the data and code, with sufficient instructions to faithfully reproduce the main experimental results.

\item {\bf Experimental setting/details}
    \item[] Question: Does the paper specify all the training and test details (e.g., data splits, hyperparameters, how they were chosen, type of optimizer) necessary to understand the results?
    \item[] Answer: \answerYes{} 
    \item[] Justification: The paper specifies all the training and test details.

\item {\bf Experiment statistical significance}
    \item[] Question: Does the paper report error bars suitably and correctly defined or other appropriate information about the statistical significance of the experiments?
    \item[] Answer: \answerYes{} 
    \item[] Justification: The paper reports appropriate information about statistical significance of the experiments.

\item {\bf Experiments compute resources}
    \item[] Question: For each experiment, does the paper provide sufficient information on the computer resources (type of compute workers, memory, time of execution) needed to reproduce the experiments?
    \item[] Answer: \answerYes{} 
    \item[] Justification: Experimental details are provided in the respective section.
    
\item {\bf Code of ethics}
    \item[] Question: Does the research conducted in the paper conform, in every respect, with the NeurIPS Code of Ethics \url{https://neurips.cc/public/EthicsGuidelines}?
    \item[] Answer: \answerYes{} 
    \item[] Justification: The research conducted in the paper conforms, in every respect, with the NeurIPS Code of Ethics.

\item {\bf Broader impacts}
    \item[] Question: Does the paper discuss both potential positive societal impacts and negative societal impacts of the work performed?
    \item[] Answer: \answerNA{} 
    \item[] Justification: Not applicable.
    
\item {\bf Safeguards}
    \item[] Question: Does the paper describe safeguards that have been put in place for responsible release of data or models that have a high risk for misuse (e.g., pre-trained language models, image generators, or scraped datasets)?
    \item[] Answer: \answerNA{} 
    \item[] Justification: Not applicable.

\item {\bf Licenses for existing assets}
    \item[] Question: Are the creators or original owners of assets (e.g., code, data, models), used in the paper, properly credited and are the license and terms of use explicitly mentioned and properly respected?
    \item[] Answer: \answerYes{} 
    \item[] Justification: The creators or original owners of assets (e.g., code, data, models), used in the paper, are properly credited and  the license and terms of use are explicitly mentioned and properly respected.

\item {\bf New assets}
    \item[] Question: Are new assets introduced in the paper well documented and is the documentation provided alongside the assets?
    \item[] Answer: \answerYes{} 
    \item[] Justification: The new assets (code and pre-trained \texttt{OTISv2} weights) will be released with documentation upon acceptance, as stated in the abstract and conclusion.

\item {\bf Crowdsourcing and research with human subjects}
    \item[] Question: For crowdsourcing experiments and research with human subjects, does the paper include the full text of instructions given to participants and screenshots, if applicable, as well as details about compensation (if any)? 
    \item[] Answer: \answerNA{} 
    \item[] Justification: Not applicable.

\item {\bf Institutional review board (IRB) approvals or equivalent for research with human subjects}
    \item[] Question: Does the paper describe potential risks incurred by study participants, whether such risks were disclosed to the subjects, and whether Institutional Review Board (IRB) approvals (or an equivalent approval/review based on the requirements of your country or institution) were obtained?
    \item[] Answer: \answerNA{} 
    \item[] Justification: Not applicable.

\item {\bf Declaration of LLM usage}
    \item[] Question: Does the paper describe the usage of LLMs if it is an important, original, or non-standard component of the core methods in this research? Note that if the LLM is used only for writing, editing, or formatting purposes and does \emph{not} impact the core methodology, scientific rigor, or originality of the research, declaration is not required.
    \item[] Answer: \answerNA{} 
    \item[] Justification: Not applicable.

\end{enumerate}

\end{document}